\documentclass[a4paper,11pt]{article}

\usepackage{gensymb}
\usepackage{fourier}
\usepackage{color}
 \usepackage{graphicx}
\usepackage{url}
\usepackage[affil-it]{authblk}
\usepackage{amsmath}
\usepackage{wrapfig}
\usepackage{blindtext}

\usepackage{lipsum,booktabs}
\usepackage{enumitem}

\usepackage{framed,multirow}

\usepackage[T1]{fontenc}
\usepackage{times}
\usepackage{multirow}
\usepackage[justification=centering]{caption}

\usepackage{amsfonts}
\newcommand{\field}[1]{\mathbb{#1}}

\newcommand{\R}{\field{R}}
\usepackage[caption=false,font=footnotesize]{subfig}

\let\OLDthebibliography\thebibliography
\renewcommand\thebibliography[1]{
  \OLDthebibliography{#1}
  \setlength{\parskip}{0pt}
  \setlength{\itemsep}{0pt plus 0.3ex}
}

\usepackage[table]{xcolor}
\usepackage{collcell}
\usepackage{hhline}
\usepackage{pgf}

\newcommand\gray{gray}

\newcommand\ColCell[1]{%
  \pgfmathparse{#1<.7?1:0}%
    \ifnum\pgfmathresult=0\relax\color{white}\fi
  \pgfmathparse{1-#1}%
  \expandafter\cellcolor\expandafter[%
    \expandafter\gray\expandafter]\expandafter{\pgfmathresult}#1}

\newcolumntype{E}{>{\collectcell\ColCell}c<{\endcollectcell}}

\DeclareMathOperator*{\argmax}{arg\,max}

\begin{document}

\title{Convolutional Neural Network on Three Orthogonal Planes for Dynamic Texture Classification}

\author{Vincent Andrearczyk \& Paul F. Whelan}
\affil{Vision Systems Group, School of Electronic Engineering, Dublin City University, Dublin 9, Ireland}
\date{}
\maketitle
\thispagestyle{empty}

\vspace{-35pt}
\begin{abstract}

Dynamic Textures (DTs) are sequences of images of moving scenes that exhibit certain stationarity properties in time such as smoke, vegetation and fire.
The analysis of DT is important for recognition, segmentation, synthesis or retrieval for a range of applications including surveillance, medical
imaging and remote sensing. 
Deep learning methods have shown impressive results and are now the new state of the art for a wide range of computer vision tasks including image and video recognition and segmentation. 
In particular, Convolutional Neural Networks (CNNs) have recently proven to be well suited  for texture analysis with a design similar to a filter bank approach.
In this paper, we develop a new approach to DT analysis based on a CNN method applied on three orthogonal planes $xy$, $xt$ and $yt$.
We train CNNs on spatial frames and temporal slices extracted from the DT sequences and combine their outputs to obtain a competitive DT classifier.
Our results on a wide range of commonly used DT classification benchmark datasets prove the robustness of our approach. Significant improvement of the state of the art is shown on the larger datasets.
\end{abstract}

\section{Introduction}
Dynamic Texture (DT) is an extension of texture in the temporal domain, introducing temporal variations such as motion and deformation.
Doretto et al. \cite{doretto2003dynamic} describe a DT as a sequence of images of moving scenes that exhibits certain stationary properties in time.
Examples of natural DTs include smoke, clouds, trees and waves. The analysis of DTs, including classification, segmentation,
synthesis and indexing for retrieval is essential for a large range of applications such as surveillance, medical image analysis and remote sensing.
New methods, mainly derived from static texture approaches are required to incorporate the analysis of temporal changes to the spatial analysis.
The major difficulties in DT analysis are due to the wide range of appearances and dynamics of DTs and to the simultaneous analysis and
combination of spatial and temporal properties.

\begin{figure*}[!t]
\centering
\includegraphics[scale=.7]{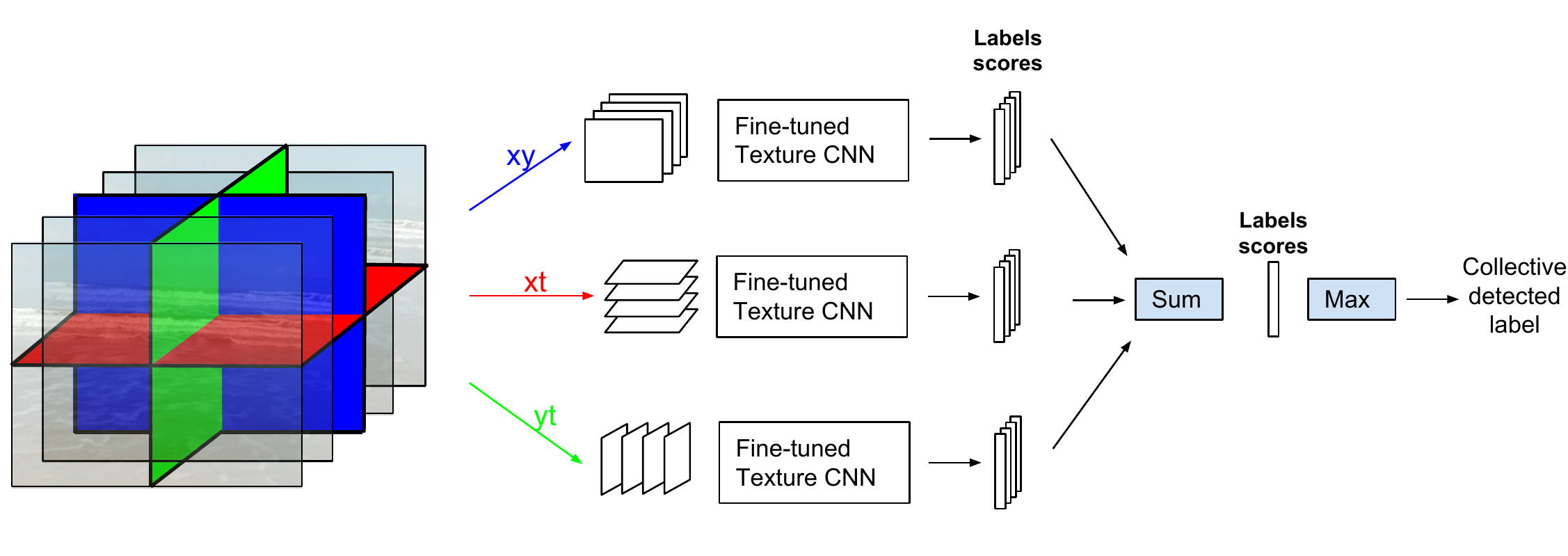}
  \caption{Overview of the proposed method for the classification of a DT sequence based on T-CNNs on three orthogonal planes in an ensemble model approach.
  The CNNs separately classify frames extracted from a input DT sequence on three planes. The outputs of the last fully-connected layers are summed and the highest score gives the collective classification decision.}\label{fig:pipeline}
\end{figure*}

Deep learning has been shown to produce state of the art results on various computer vision tasks. In particular, Convolutional Neural Networks (CNNs) are very powerful for image segmentation and classification 
and have shown great adaptability to texture analysis by discarding the overall shape analysis \cite{andrearczyk2016using,cimpoi2014deep,lin2015visualizing}.
This paper presents a new CNN framework for the analysis of DTs and an application to DT classification. 
The idea of our approach is to extend our previous work on Texture CNN (T-CNN) \cite{andrearczyk2016using} to the analysis of DTs by analyzing the sequences on three orthogonal planes. 
Thus we train neural networks to recognize DT sequences based on the spatial distribution of pixels and on their evolution and dynamics over time.
This is partly inspired by the extension of Local Binary Pattern on Three Orthogonal Planes (LBP-TOP) by Zhao and Pietik\"ainen \cite{zhao2007dynamic}.
An overview of the pipeline of the proposed method is illustrated in Figure \ref{fig:pipeline} and explained in more detail in Section \ref{sec:Methods}.
We first extract slices from the DT sequences to train (i.e. from scratch or fine-tune) an independent T-CNN on each plane.
We then sum the outputs of all the slices on all the planes to obtain the class with maximum prediction score during the testing phase in an ensemble model fashion \cite{hansen1990neural}. 
We test our method on widely used DT classification datasets and compare our results to the state of the art.

The main contributions of this paper are as follows:
\\(a) We introduce a new framework to analyze DTs on three orthogonal planes.
\\(b) We develop deeper CNNs for textures as well as networks adapted to small images based on the original T-CNN in \cite{andrearczyk2016using}.
\\(c) We experiment on three DT databases with seven benchmarks and present comparative results with the literature.
\\(d) We conduct an analysis of the contribution and complementarity of each plane in the learning and classification of DT sequences as well as the domain transferability of trained parameters.

The rest of the paper is organized as follows. In Section \ref{sec:RelWork}, we describe the related work. We present our CNN method for DT analysis in Section \ref{sec:Methods}.
In Section \ref{sec:Exp}, we test our method on three widely used DT databases derived in seven benchmarks with considerable differences including the number of classes, number and size of images and the experimental protocol.
We compare our results to the state of the art and show a significant improvement.
\section{Related work}
\label{sec:RelWork}

\subsection{Deep learning for texture images}
Basu et. al \cite{basu2016theoretical} argue that the dimensionality of texture datasets is too large for deep networks to shatter them without explicitly extracting hand-crafted features beforehand.
However, \cite{andrearczyk2016using,cimpoi2014deep,lin2015visualizing} show that the domain transferability of deep networks enables pre-trained networks to perform very well on texture classification and segmentation tasks as a simple feature extractor \cite{cimpoi2014deep} or in an end-to-end training and testing scheme \cite{andrearczyk2016using,lin2015visualizing}.
As explained in \cite{andrearczyk2016using}, a CNN approach is well suited to texture analysis as it can densely pool texture features.
Indeed, features learned by the first layer are similar to Gabor filters with mostly edge-like kernels and deeper features can be used as a more complex filter bank approach, widely used in texture analysis \cite{gonccalves2012spatiotemporal,randen1999filtering}.

In \cite{sifre2013rotation}, Siffre and Mallat derive texture descriptors that are extracted in a similar manner as a CNN approach.
The trainable filters are replaced by hand-crafted scaled and rotated wavelets to perform a scattering transform.
The obtained descriptors (in the scattering domain) are invariant to transformations such as translation, scaling and rotation and robust to small deformations.
Rotation invariance is developed in a shallow CNN for texture classification in \cite{marcos2016learning} by tying the weights of multiple rotated versions of filters.

Deep neural networks were applied to texture analysis in \cite{cimpoi2014deep} as a feature extractor,
in which the output of the last convolution layer is used in a framework including PCA and SVM classifier. They largely improve the state of the art on texture classification benchmarks with very deep networks.
A bilinear CNN model is developed in \cite{lin2015bilinear} for fine-grained recognition, combining two CNNs to extract and classify translationally invariant local pairwise features in a neural network framework.
They are also able to generalize several widely used orderless texture descriptors. This approach also performs well on texture datasets as shown in \cite{lin2015visualizing}.
A T-CNN is developed in \cite{andrearczyk2016using} which includes an energy layer that extracts the dense response to intermediate features in the network, obtaining significant results on texture classification tasks with reduced complexity compared to classic CNNs.
The complementarity of texture and shape analysis is shown using convolutional networks for feature extraction in \cite{cimpoi2014deep} and in an end-to-end CNN training scheme in \cite{andrearczyk2016using}.
A fully convolutional approach is developed for the segmentation of texture regions in \cite{andrearczyk2017texture}.

\subsection{Classic Dynamic Texture approaches}
In this section we present methods mainly developed for the analysis of DTs though some are applied to the close field of dynamic scene analysis.
Difficulties in DT analysis arise from the large range of phenomena resulting from natural scenes and recording methods including scale, illumination and rotation variation as well as static and moving cameras.
Most classic DT analysis approaches that attempt to overcome all or some of these difficulties can be classified into four categories, namely statistical, motion based, model based and spatiotemporal filtering.

\textit{Statistical} approaches \cite{zhao2007dynamic,flores2005identifying,tiwari2016improved,tiwari2016dynamic,rahtu2012local,chen2013automatic,yang2016dynamic} mainly adapt standard spatial texture methods such as Grey Level Co-occurrence matrix (GLCM) \cite{haralick1973textural} and Local Binary Pattern (LBP) \cite{ojala1996comparative} to extend their analysis to the spatio-temporal domain. 
LBP has been widely investigated and developed in both static and dynamic texture analysis due to its computation simplicity, invariances and good performance. 
Zhao and Pietik\"ainen \cite{zhao2007dynamic} extend the Local Binary Pattern (LBP) to DTs by extracting LBPs on Three Orthogonal Planes (LBP-TOP):
$xy$ which is the classic spatial texture LBP as well as $xt$ and $yt$ which consider temporal variations of the pixel intensities.
Several approaches were derived from the LBP-TOP. The Local Phase Quantization on Three Orthogonal Planes (LPQ-TOP) \cite{rahtu2012local} based on
quantizing the phase information of the local Fourier transform which is particularly robust to blurred images. 
The LBP-TOP is combined to a Weber Local Descriptor on Three Orthogonal Planes in \cite{chen2013automatic} to describe the spatial appearance of DTs for segmentation, combined to a motion analysis for the temporal information.
More recently, Tiwari and Tyagi \cite{tiwari2016improved} incorporate a noise resistance feature to the LBP-TOP with a Weber's law
threshold in the creation of the patterns and achieve high accuracy on the Dyntex++ \cite{ghanem2010maximum} and UCLA \cite{doretto2003dynamic} databases.
The main drawbacks of the LBP-TOP approaches are the poor robustness to camera motion and the limited amount of information that it is able to extract with the hand-crafted patterns.

\textit{Motion based} methods exploit the statistical properties of the extracted motion between consecutive frames of the DT sequences.
Several motion extraction methods are used for the analysis of DT including Local Motion Pattern (LMP) \cite{gao2008extended}, Complete Flow (CF) \cite{chen2013automatic,andrearczyk2015dynamic} and more commonly Normal Flow (NF) \cite{peh2002synergizing,peteri2005dynamic,fazekas2005normal,fazekas2007dynamic,lu2005dynamic,bouthemy1998motion}.
Statistical features are then extracted from the motion to describe its spatial distribution such as GLCM \cite{andrearczyk2015dynamic,peh2002synergizing,bouthemy1998motion},
Fourier spectrum \cite{peh2002synergizing}, difference statistics \cite{peh2002synergizing}, histograms \cite{chen2013automatic,gao2008extended,lu2005dynamic}
and other statistics calculated on the motion field itself \cite{peteri2005dynamic} and on its derivatives \cite{fazekas2007dynamic}.

\textit{Model based} methods \cite{doretto2003dynamic,saisan2001dynamic,ravichandran2013categorizing,chaudhry2013dynamic,afsari2012group,fujita2003recognition}
aim at estimating the parameters of a Linear Dynamical System (LDS) using a system identification theory in order to capture the spatial appearance and dynamics of a scene.
Originally designed for the synthesis problem, the estimated parameters can be used for a classification task \cite{doretto2003dynamic}.
Positive results were obtained in the learning and synthesis of temporal stationary sequences such as waves and clouds in \cite{doretto2003dynamic}.
However, the model based approach raises several difficulties such as the distance between models lying in a non-linear space of LDSs with a complicated manifold as well as a poor
invariance to rotation, scale and illumination and assumes a single, well segmented DT.
This approach is combined to a kernel based tracking framework in \cite{chaudhry2013dynamic} for tracking DTs in videos.
In \cite{fujita2003recognition} Fujita and Nayar bypass the non-linear space, segmentation and presence of multiple DTs problems by using impulse response of state variables learned during the system identification which capture the fundamental dynamical properties of DTs.
Ravichandran et al. \cite{ravichandran2013categorizing} overcome the view-invariance problem using a Bag-of-dynamical Systems (BoS) similar to a Bag-of-Features (BoF) with LDSs as feature descriptors.
Afsari et. al \cite{afsari2012group} define a new distance in the non-linear space between LDSs based on the notion of aligning LDSs.
It enables them to compute the mean of a set of LDSs for an efficient classification, particularly adapted to large-scale problems.

\textit{Filtering} approaches for texture analysis were extended to the spatiotemporal domain for the analysis of DTs \cite{derpanis2010dynamic,yu2011dynamic,gonccalves2012spatiotemporal,ji2013wavelet,feichtenhofer2014bags}. 
Derpanis and Wildes \cite{derpanis2010dynamic} use spacetime oriented filters to extract interesting features which describe intrinsic properties of the DTs
such as unstructured, static motion and transparency.
In \cite{yu2011dynamic}, Qiao and Wang use 3D Dual Tree Complex Wavelet combining spatial and dynamic analyses.
Other approaches use Wavelet Domain Multifractal Analysis (WDMA) \cite{ji2013wavelet}, Spatiotemporal Oriented Energy (SOE) \cite{feichtenhofer2014bags} and spatiotemporal Gabor filters \cite{gonccalves2012spatiotemporal}.

Apart from these four major categories of DT approaches, other methods have also attracted interest such as the graph-based approach used in \cite{ramirez2015spatiotemporal} in which graph-based descriptors are extracted using 3D filters to represent the signature of DTs for a classification scheme.

Finally, many approaches combine two or more approaches for their complementarity in the spatiotemporal analysis \cite{chen2013automatic,yang2016dynamic,peteri2005dynamic,andrearczyk2015dynamic,gao2008extended}.
In \cite{ghanem2010maximum}, Ghanem and Ahuja combine LBP, Pyramid of Histogram of Oriented Gradients and an LDS model approach to jointly analyze the spatial texture, the spatial layout and the dynamics of DT sequences.
More recently, Yang et. al \cite{yang2016dynamic} make use of ensemble SVM to combine static features such as LBP and responses to Gabor filter banks with temporal model based features.

\subsection{Deep learning for videos and Dynamic Textures}
Following the breakthrough of deep learning in image recognition, several attempts have been made to apply CNNs to videos and/or DT analysis.

End-to-end convolutional networks with 3D filters are developed in \cite{ji20133d}, capturing spatial and temporal information for human action recognition.
In \cite{tran2014learning} Tran et. al generalize the 3D convolution and pooling to cope with large video datasets.
They obtain excellent results on action, scene and object recognition.
In \cite{karpathy2014large}, several CNN architectures are evaluated on a video classification task including classic single-frame CNN, two-stream single frame networks with various fusion approaches and 3D convolution filters.
However, their attempt to combine motion analysis to the classic spatial convolutional network by using stacks of images as input results in a surprisingly modest improvement.
While the analysis of a few sequences benefit from the motion analysis, the latter increases the sensitivity to camera motion.

Simonyan and Zisserman \cite{simonyan2014two} develop a two-stream CNN method including spatial and temporal analysis for action recognition.
In their architecture, a spatial network is trained to recognize still video frames while a motion network is trained on dense optical flow.
The two networks are trained separately and their softmax scores are combined by late fusion with a linear SVM and, with lower accuracy, by averaging. With their approach they demonstrate the complementarity of spatial and temporal analysis in video classification.
Donahue et al. \cite{donahue2015long} develop a long-term recurrent CNN to learn compositional representations in space and time for video analysis.
This recurrent approach is able to learn complex temporal information from sequential data on account of a temporal recursion.
While this method is appropriate for activity recognition, image captioning, and video description, it is not as relevant for DT analysis in which one is more interested in probabilities on the distribution of
the pixel intensities (in time and space) than in the sequential detection of particular events.

Shao et al. \cite{shao2016slicing} use the semantic selectiveness of spatial convolutional filters to discard the background in crowd scenes followed by a combination of
spatial and temporal convolutions in a CNN architecture for crowd video understanding.

In \cite{shaocrowded}, the same authors use a fully convolutional network on spatial and temporal slices and on the motion extracted on theses slices
with various fusion strategies in order to create a per-pixel collectiveness map from crowd videos. 
While these approaches share some similarities with our proposed method, they are designed for the analysis of crowded scenes
with assumptions (e.g. background) and methods (e.g. collectiveness map) which do not apply to DT recognition.
To the best of our knowledge, these two approaches are the only ones using spatial and temporal slices in a convolutional network approach.

While the deep learning approaches for video analysis introduced so far were mainly developed for human action recognition, several recent neural network approaches have focused on classifying DTs.
Culibrk and Sebe \cite{culibrk2014temporal} use temporal gradients and Group Of Pictures (GOP)
as inputs to CNNs to classify DTs. However, in order to use the pre-trained AlexNet, the authors dilate and warp the $50\times 50$ images to the AlexNet input size $227\times 227$.
This pre-processing step results in a waste of computation and a network architecture which is not adapted to the input images.
Similar to the spatial approach in \cite{cimpoi2014deep}, in which CNNs are used to extract features from static texture images without learning from the texture database, Qi et. al 
\cite{qi2016dynamic} use a CNN to extract features from the DT database. They apply a pre-trained CNN as a feature extractor to obtain mid-level features from each
frame of the sequence and then compute and classify the first and second order statistics. They obtain the state of the art results on the classification of the DynTex database
(alpha, beta and gamma benchmarks) \cite{peteri2010dyntex}.
Finally in \cite{yan2014modeling}, a variant of stacked autoencoder is trained to learn video dynamics which can be more complex than those based on LDS estimation.
The deep encoded dynamics can be used similarly to classic model based methods to synthesize and recognize DTs and segment motions.

\subsection{Ensemble models}
An ensemble model is a supervised learning method that combines a set of classifiers to create a more accurate one by averaging their predictions.
This approach is mostly used to avoid underfitting (e.g., boosting) or overfitting (e.g., bagging).
The most popular approaches used to combine predictions from multiple models are grouped into algebraic and voting methods.
Algebraic methods compute basic statistics to combine the predictions such as average, sum and median, with a possible weighting scheme to emphasize the most accurate classifiers' predictions in the collective decision.
Voting methods include the majority which chooses the classification made by more than half the networks and the borda count \cite{ho1994decision} based on ranking votes of the model classifiers.
The benefit of ensemble models for neural networks was revealed in \cite{hansen1990neural}.
Training neural networks is difficult and depends on the initialization of the weights and biases, the data splits and is highly subject to overfitting.
An ensemble of independently trained networks can improve the predictions by reducing the overfit and avoiding the possible poor test result of a single network.

The basic idea of training a set of deep neural networks on the same data or resampled versions of it to avoid overfitting was outperformed by new methods such as
dropout \cite{srivastava2014dropout}, dropconnect \cite{wan2013regularization} or stochastic pooling \cite{zeiler2013stochastic} layers
which achieve similar behaviors and better results with a single network and therefore a much lower complexity.
However, in a data fusion ensemble model, multiple networks are necessary to analyze heterogeneous input data \cite{ahmad2005combination,yim1995neural}.
In other words, independent networks learn different data modalities (e.g., multiple sensors, images, videos, depth and audio) to make a collective classification decision.

In this paper we use a data (late) fusion ensemble model as we analyze three planes which represent three data modalities.
We use a sum scheme to combine the models as the confidence given by the output of the CNNs is valuable and would be discarded using a voting method such as majority or borda count.
The sum score also showed better results experimentally. Additional details on the ensemble model method used in this paper are provided in Section \ref{subsec:DTCNN}.
Note that this ensemble model approach is often referred to as ``late fusion" to combine the output of separately trained networks \cite{karpathy2014large,simonyan2014two,shao2016slicing}.

\section{Methods}
\label{sec:Methods}
This section describes our CNN approach for DT analysis and presents a new network adapted to small input image sizes as well as deep architectures based on \cite{andrearczyk2016using}.
\subsection{Texture CNN}
\label{subsec:TCNN}
The Texture-CNN (T-CNN) approach\footnote{An implementation of the T-CNN-3 to train and test on the kth-tips-2b texture database is available here:\\https://github.com/v-andrearczyk/caffe-TCNN} was introduced in our previous work \cite{andrearczyk2016using} and applied to a tissue classification task in \cite{andrearczyk2016deep}.
T-CNN is a network of reduced complexity adapted to the analysis of texture images by making use of dense features extracted at intermediate layers of the CNN and discarding the overall shape analysis of a classic convolutional network by introducing an energy layer.
The energy layer pools the average response of all the feature maps into a vector of size equal to the number of input feature maps.
\\The vector output $\mathbf{e} \in \R^K$ of an energy layer is calculated as:
\begin{equation} \label{eq:energy}
\mathbf{e}[k] = \dfrac{1}{N\times M}\sum\limits_{i=0}^N\sum\limits_{j=0}^M x_{i,j}^k ,
\end{equation}
where $x_{i,j}^k$ is the value of the $k^{th}$ input feature map at spatial location $(i,j)$. $k$ enumerates the $K$ input feature maps and $N$, $M$ are respectively the number of rows and columns of the input feature maps.
For fixed square input sizes, this layer can be implemented with an average pooling layer of kernel size equal to the input feature map size. 

\begin{figure*}[!t]
\centering
\includegraphics[scale=.6]{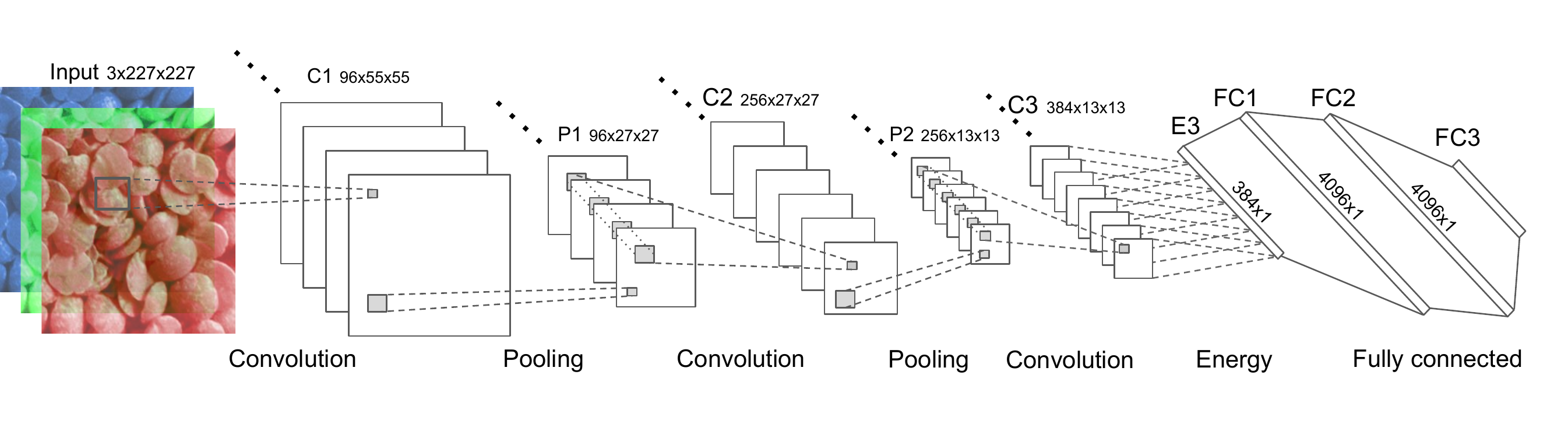}
  \caption{Texture CNN architecture using three convolution layers (T-CNN-3) \cite{andrearczyk2016using}.}\label{fig:TCNN_color_C3}
\end{figure*}

This approach achieves better results than a classic CNN (AlexNet), using the same architecture for the first and final layers and with a number of trainable parameters nearly three times smaller.
The architecture of the T-CNN-3 (with three convolution layers) used in this paper for the DynTex database is depicted in Figure \ref{fig:TCNN_color_C3}.
More details on the architecture are provided in Table \ref{tab:T-CNN-3} and in \cite{andrearczyk2016using}.
Another network based on the same textural approach is developed to enable the analysis of smaller input sizes for the Dyntex++ and UCLA datasets as described below.
In this paper, we experiment on a deeper T-CNN approach based on GoogleNet \cite{szegedy2015going}.
The same idea of energy layer is used after the inception 4a block.

\subsection{Small Texture CNN (T-CNN50)}
\label{subsec:SmallCNN}
A CNN is used in \cite{culibrk2014temporal} to analyze DT sequences of the Dyntex++ database. The authors dilate and warp the input images in order to match the input size of AlexNet ($227\times 227$). 
This approach is not optimal both in terms of complexity of the network and capability to train and learn from the small images.

Instead, we have designed a new CNN architecture based on T-CNN described previously to analyze small input images such as those of the Dyntex++ sequences ($50\times 50$) and UCLA sequences ($48\times 48$). 
While the input size of the T-CNN does not need to be fixed due to the energy layer, it is necessary to develop another architecture for these small input sizes for the following reasons.
We want to keep the receptive field of the neurons small enough to tile the input image in such a way that the energy layer will behave similarly to a filter bank that spans the input image.
If we kept the architecture of the T-CNN, the effective receptive field of the neurons of the third convolution layer would be larger than the input image itself.
The architecture of this new network (that we refer to as \textit{T-CNN50} in this paper) is detailed in Table \ref{tab:T-CNN50}.
Similarly to the analysis of larger images, we develop a deeper network based on GoogleNet and T-CNN by reducing the receptive fields while maintaining the depth.

An important aspect of CNNs is the domain transferability e.g. pre-training a network on a very large database and fine-tuning on a smaller target dataset shows that the pre-trained features generalize well to new types of data in different domains \cite{yosinski2014transferable}.
Also, one can generally transfer some or all the kernels learned by any network to another network if the kernel sizes are equal, including a CNN with smaller input size.
However, the kernel sizes of the T-CNN and the T-CNN50 being different, we cannot initialize the T-CNN50 using the kernels learned by the T-CNN on ImageNet.
Therefore we pre-train the T-CNN50 (based on AlexNet and GoogleNet) on ImageNet \cite{deng2009imagenet} with the images resized to $50\times 50$.
\begin{table}[!t]
\caption{Architecture of the T-CNN-3 \cite{andrearczyk2016using}, where c is the number of color channels and N the number of classes.} \label{tab:T-CNN-3}
\centering
\resizebox{0.6\textwidth}{!}{
\begin{tabular}{ |c|c|c|c| }
\hline
	Layer type & output size & kernel, pad, stride & train. param. \\ \hline
    crop & $c\times 227\times 227$ & - & 0 \\ \hline
	Conv (C1) & $96\times 55\times 55$ & 11, 0, 4 & $c\times 11616 + 96$ \\ \hline
	ReLU & $96\times 55\times 55$ & - & 0 \\ \hline
	Pool (P1) & $96\times 27\times 27$ & 3, 0, 2 & 0 \\ \hline
	LRN & $96\times 27\times 27$ & -  & 0 \\ \hline
	Conv (C2) & $256\times 27\times 27$ & 5, 2, 1 & 614,656 \\ \hline
	ReLU & $256\times 27\times 27$ & - & 0 \\ \hline
	Pool (P2) & $256\times 13\times 13$ & 3, 0, 2 & 0 \\ \hline
	LRN & $256\times 13\times 13$ & - & 0 \\ \hline
	Conv (C3) & $384\times 13\times 13$ & 3, 1, 1 & 885,120 \\ \hline
	ReLU & $384\times 13\times 13$ & - & 0 \\ \hline
	Energy & $384$ & - & 0 \\ \hline
	Fully-con. (FC1) & 4096 & - & 1,576,960 \\ \hline
	ReLU & 4096 & - & 0 \\ \hline
	Dropout & 4096 & - & 0 \\ \hline
	Fully-con. (FC2) & 4096 & - & 16,781,312 \\ \hline
	ReLU & 4096 & - & 0 \\ \hline
	Dropout & 4096 & - & 0 \\ \hline
	Fully-con. (FC3) & N & - & $4096\times N+N$ \\ \hline
	Softmax & N & - & 0 \\ \hline
    
\end{tabular}
}
\end{table}
\begin{table}[!t]
\caption{Architecture of the proposed T-CNN50 . Where c is the number of color channels and N the number of classes.} \label{tab:T-CNN50}
\centering
\resizebox{0.6\textwidth}{!}{
\begin{tabular}{ |c|c|c|c| }
\hline
	Layer type & output size & kernel, pad, stride & train. param. \\ \hline
    crop & $c\times 48\times 48$ & - & 0 \\ \hline
	Conv (C1) & $96\times 48\times 48$ & 5, 2, 1 & $c\times 2400+96$ \\ \hline
	ReLU & $96\times 48\times 48$ & - & 0 \\ \hline
	Pool (P1) & $96\times 24\times 24$ & 2, 0, 2 & 0 \\ \hline
	LRN & $96\times 24\times 24$ & -  & 0 \\ \hline
	Conv (C2) & $256\times 24\times 24$ & 3, 1, 1 & 221,440 \\ \hline
	ReLU & $256\times 24\times 24$ & - & 0 \\ \hline
	Pool (P2) & $256\times 12\times 12$ & 2, 0, 2 & 0 \\ \hline
	LRN & $256\times 12\times 12$ & - & 0 \\ \hline
	Conv (C3) & $384\times 12\times 12$ & 3, 1, 1 & 885,120 \\ \hline
	ReLU & $384\times 12\times 12$ & - & 0 \\ \hline
	Energy & $384$ & - & 0 \\ \hline
	Fully-con. (FC1) & 3000 & - & 1,155,000 \\ \hline
	ReLU & 3000 & - & 0 \\ \hline
	Dropout & 3000 & - & 0 \\ \hline
	Fully-con. (FC2) & 3000 & - & 9,003,000 \\ \hline
	ReLU & 3000 & - & 0 \\ \hline
	Dropout & 3000 & - & 0 \\ \hline
	Fully-con. (FC3) & N & - & $3000\times N+N$ \\ \hline
	Softmax & N & - & 0 \\ \hline
    
\end{tabular}
}
\end{table}

\subsection{Dynamic Texture CNN}
\label{subsec:DTCNN}
The main idea of our approach is to use CNNs on three orthogonal planes.
Our method thus learns and pools dense features that are expected to be repetitive in the spatial and temporal domains.
This approach is partly inspired by the LBP-TOP \cite{zhao2007dynamic} which extracts LBP patterns on three orthogonal planes.
It is also influenced by the ensemble models \cite{hansen1990neural} which extract useful diverse knowledge from the training data by learning
different models in parallel and averaging their prediction.
The overall framework of our Dynamic Texture CNN (DT-CNN) approach is detailed in Figure \ref{fig:pipeline}.
We refer to the developed DT recognition frameworks as DT-AlexNet and DT-GoogleNet depending on which T-CNN architecture is used.

\subsubsection{Slicing the Dynamic Texture data}\leavevmode \par
\begin{figure*}[!t]
\centering
\includegraphics[scale=.72]{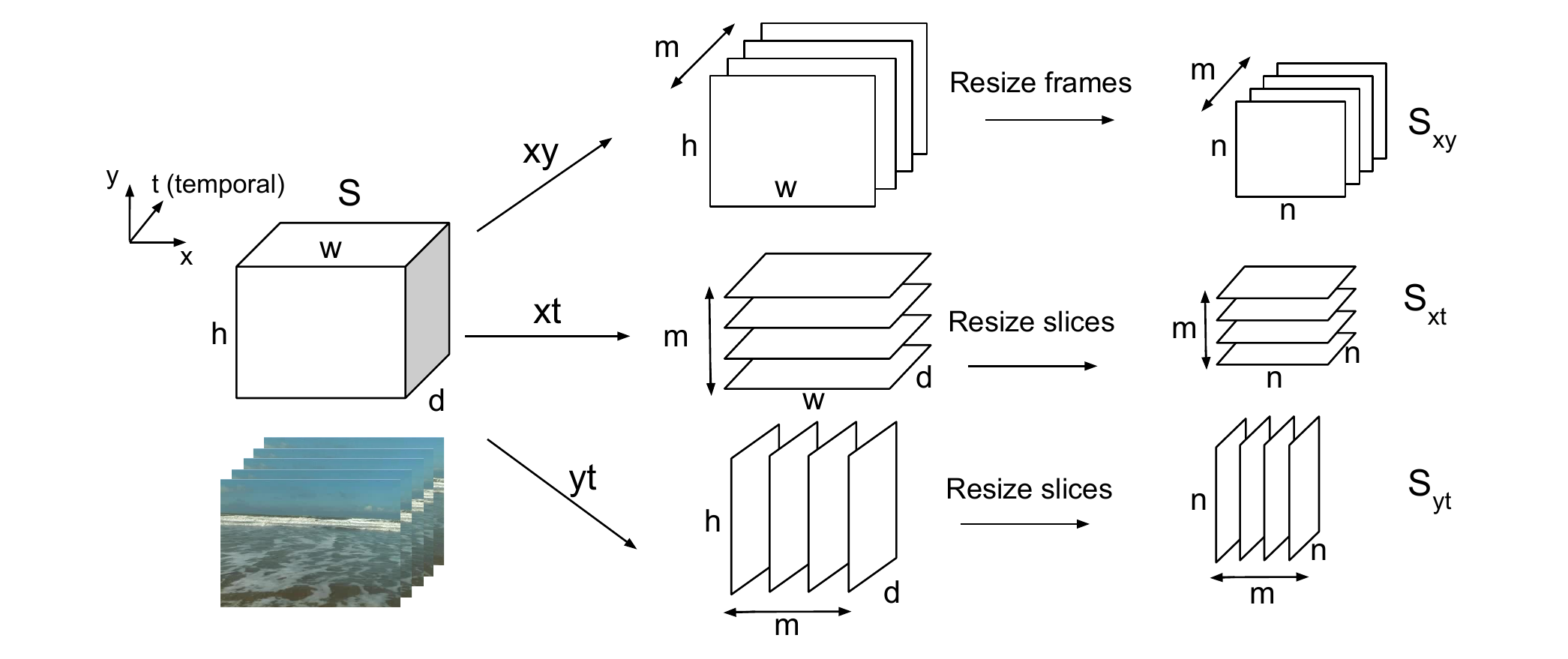}
  \caption{Diagram of DT sequence slicing in three orthogonal planes.}\label{fig:xyt_sequences}
\end{figure*}

We must process (slice) the sequences of DT to enable the CNNs to train and test on the three orthogonal planes. The diagram explaining the pre-processing of a DT sequence is shown in Figure \ref{fig:xyt_sequences}.

\textit{XY plane (spatial):}
We represent a sequence of DT with $d$ frames of size $h\times w$ as a matrix \textbf{$S$} $\in \R^{h\times w\times d\times c}$ with $h$ (height), $w$ (width) and $d$ (depth) respectively in the $x$, $y$ and $t$ axes and $c$ the number of color channels (i.e., three for rgb, one for greyscale).
In the spatial plane, we simply extract $m$ slices (frames) equally spaced in the temporal axis from \textbf{$S$}. We resize all the frames using bilinear interpolation to the size $n\times n$ to obtain a sequence \textbf{$S_{xy}$} $\in \R^{n\times n\times m\times c}$  with $m \leq  \min(d,h,w)$ and $n \leq \min(d,h,w)$.

\textit{XT plane (temporal):}
From the same sequence \textbf{$S$}, we extract $m$ slices in the $xt$ plane, equally spaced on the $y$ axis. We resize the slices to the size $n\times n$ to obtain a sequence \textbf{$S_{xt}$} $\in \R^{n\times n\times m\times c}$ with $m \leq \min(d,h,w)$ and $n \leq \min(d,h,w)$.
A slice in the $xt$ plane reflects the evolution of a row of pixels over time throughout the sequence.

\textit{YT plane (temporal):}
Finally, we extract $m$ slices in the $yt$ plane, equally spaced on the $x$ axis. We resize the slices to the size $n\times n$ to obtain a sequence \textbf{$S_{yt}$} $\in \R^{n\times n\times m\times c}$ with $m \leq \min(d,h,w)$ and $n \leq \min(d,h,w)$.
A slice in the $yt$ plane reflects the evolution of a column of pixels over time throughout the sequence.
\\After pre-processing the sequences, we obtain three sets of sequences of slices which represent the same sequences in three different planes.
Examples of spatial and temporal slices are shown in Figure \ref{fig:fig_slices}.

\subsubsection{Training on three planes}\leavevmode \par
In order to train the CNNs, we split the sequences into training and testing sets.
Details on the training and testing splits are provided in the experimental setups in Section \ref{subsec:Datasets}.
A dataset containing $M$ original sequences is split into $N$ training sequences and ($M-N$) testing sequences. In each plane, we have $N\times m$ training and $(M-N)\times m$ testing slices, with $m$ the number of slices per sequence.
In each plane, the training slices extracted from it are used to fine-tune an independent neural network. In the testing phase, we classify all the slices in each plane and combine the outputs as explained in the following section.
The networks used in our experiments are detailed in Sections \ref{subsec:TCNN} and \ref{subsec:SmallCNN} while the hyperparameters are described in Section \ref{subsec:ImpDet}.

\subsubsection{Sum collective score}\leavevmode \par
Because we train an independent network for each of the three orthogonal planes, we must combine the outputs in the testing phase.
We operate a collective score by summing the outputs of the CNNs. 
First, since each sequence is represented in each plane by a stack of slices, we obtain the score in each plane for a particular sequence by summing the outputs of all the slices of a same sequence.

If we have $m$ slices per sequence, the score vector of a sequence in a plane $p$ is given as:
\begin{equation} \label{eq:score1}
\mathbf{s}^p = \sum\limits_{i=1}^m \mathbf{s}_{i}^p ,
\end{equation}
\\where $\mathbf{s}_{i}^p \in \R^N$ is the output (non-normalized classification score) of the last fully-connected layer of the $i^{th}$ slice of the input sequence on plane $p$, with $p \in \{xy, xt, yt\}$.
$N$ is the number of classes. Note that all the scores $\mathbf{s}_{i}^p$ with $i =\{1,...,N\}$ are obtained with the same fine-tuned network for a particular plane $p$ and that each plane uses an independently fine-tuned network.
\\We then obtain a global score for a particular sequence by summing over the three planes.
\begin{equation} \label{eq:score2}
\mathbf{s} = \sum\limits_{p=\{xy,xt,yt\}} \mathbf{s}^p
\end{equation}
Note that in the experiments we also sum over two planes or single planes to analyze their contribution and complementarity.
\\The collectively detected label $l$ for a sequence is the one for which the sum score $\mathbf{s}$ is maximum:
\begin{equation} \label{eq:score3}
l = \argmax\limits_{i}(\mathbf{s}[i]) ,
\end{equation}
where $i =\{1,...,N\}$ enumerates the DT classes.

This ensemble model approach combines three weak neural network classifiers to create a more accurate one as suggested in \cite{hansen1990neural}.
We use a data fusion approach as it requires three network classifiers to recognize three data types derived from the sequences (i.e. sequence slices in three different planes). 
By using a sum collective score, we take into consideration the classification confidence given by the output vector of the last fully-connected layer.
We define confidence in this context as the magnitude of the output neurons of the convolutional network. Each neuron gives a score similar to a non-normalized probability for the input image to belong to a certain class.

Also, the confidence sum of the raw output of the last fully-connected layer gives better results than the sum of the softmax normalized probability output.
Using the raw output, large non-normalized scores can be attributed to a sequence by a single plane for a particular class if the confidence is high.
This is similar to an automatic weighting strategy based on the detection confidence of each networks.

We also confirmed experimentally that the results obtained with this sum collective score are better than using a majority or a borda count voting scheme.

\subsection{Domain Transfer}
We use CNNs pre-trained on ImageNet to transfer the knowledge learned from this large image dataset to our three planes analysis.
While we are only able to pre-train on the spatial ($xy$) axis since ImageNet does not contain videos, we find out that it transfers relatively well to the analysis of temporal axes ($xt$ and $yt$).
As suggested in \cite{yosinski2014transferable}, transferring features from the relatively distant task of spatial images recognition is better than using random features for the recognition of temporal images.
Although the improvement in terms of accuracy is minor, using pre-trained networks also makes the training significantly faster.

Finally, we are aware of existing video datasets such as UCF101 (human actions \cite{soomro2012ucf101}) and Sports-1M \cite{karpathy2014large}.
However, our approach is designed for the classification of segmented DTs. The temporal slices are extracted at multiple spatial locations of the sequence and should exhibit the dynamic of the analyzed DT.
Therefore, we do not pre-train our networks on such video datasets as slicing such sequences would extract multiple types of dynamic with potentially only a few slices that represent the dynamic of interest.
\begin{figure}[!t]
\centerline{\subfloat[xy]{\includegraphics[width=1in]{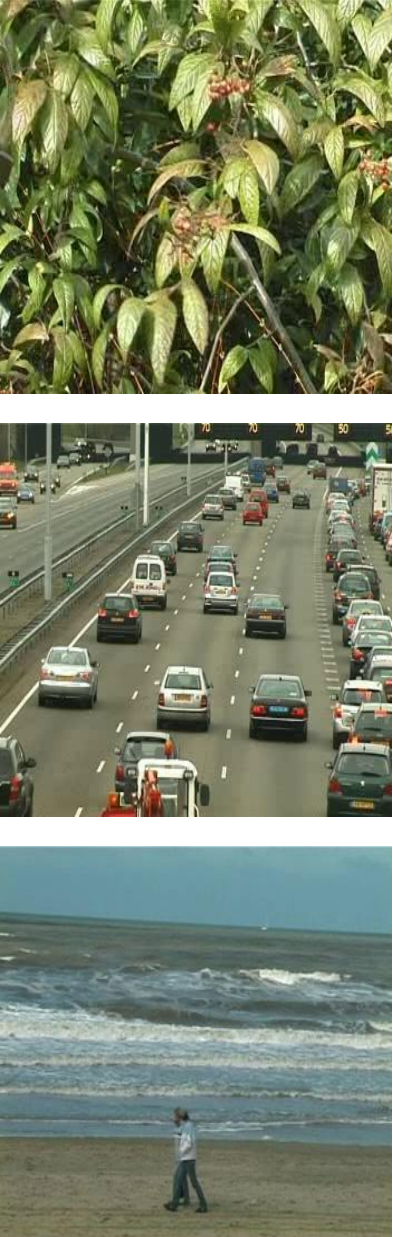}%
\label{fig:slicesxy}}
\hfil
\subfloat[xt]{\includegraphics[width=1in]{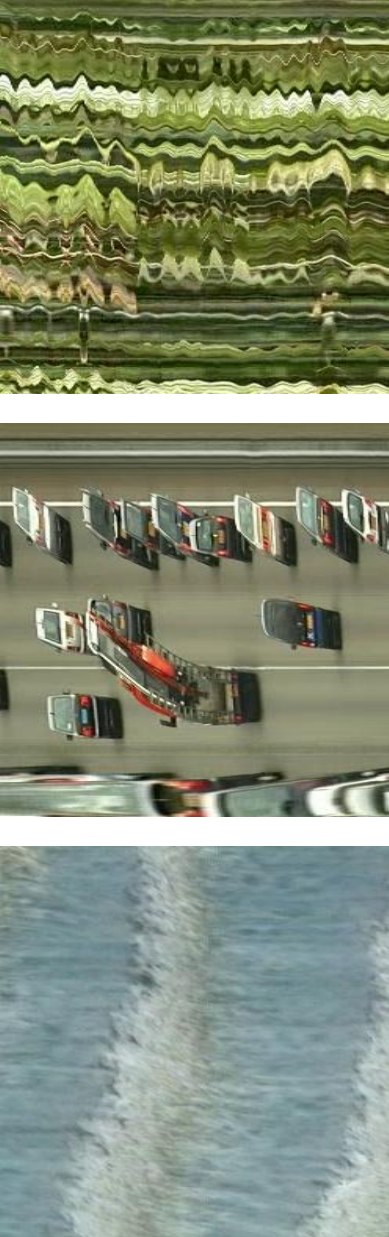}%
\label{fig:slicesxt}}
\hfil
\subfloat[yt]{\includegraphics[width=1in]{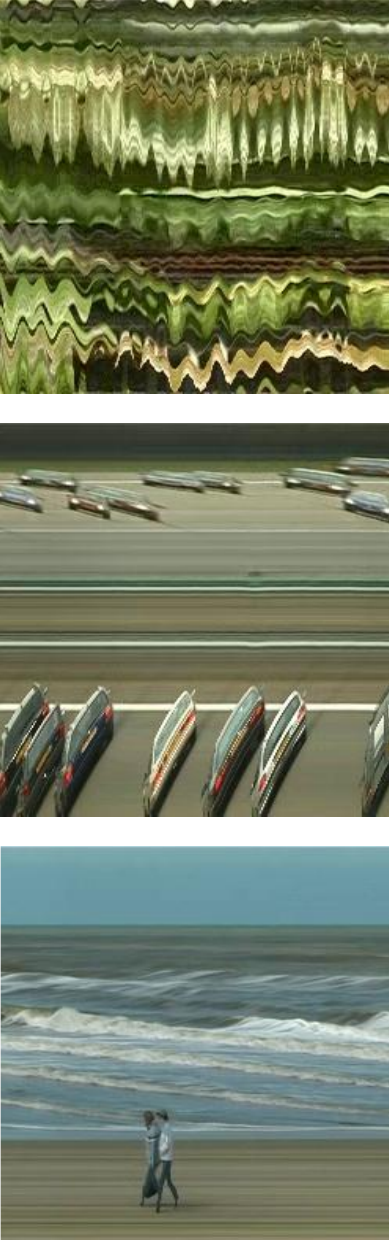}%
\label{fig:slicesyt}}}
\caption{Examples of DT slices in three orthogonal planes of foliage, traffic and sea sequences from the DynTex database. (a) xy (spatial), (b) xt (temporal) and (c) yt (temporal).}
\label{fig:fig_slices}
\end{figure}

\section{Experiments}
\label{sec:Exp}
In this section, we present the datasets and protocols used in our experiments and provide implementation details.
The results obtained with our approach are compared with the state of the art.

\subsection{Datasets}
\label{subsec:Datasets}
In our experiments, we use three datasets organized in seven benchmarks to test our algorithm and compare our results to the state of the art.
These experiments cover most of the datasets used in the literature.
They include many DT types, various number of training samples, balanced and unbalanced classes, static and moving cameras, rotation, illumination and scale variation as well as several validation setups (leave-one-out, N-folds cross-validation and random splits).

The DynTex Database \cite{peteri2010dyntex} is a diverse collection of high-quality DT videos.
Three sub-datasets are compiled from the DynTex database. For all of them, we use the processed and downsampled ($352\times 288$) color images.
All sequences are at least 250 frames long, we therefore use the first 250 frames of all sequences to extract the slices. 
We use the same experimental setup as in \cite{qi2016dynamic} with a leave-one-out classification.
The three sub-dataset are defined as follows.

\textit{DynTex alpha} contains 60 DT sequences equally divided into three categories: ``sea" (20), ``grass" (20) and ``trees" (20), where the numbers represent the number of sequences in each class.

\textit{DynTex beta} contains 162 sequences divided into 10 classes:
``sea'' (20), ``vegetation" (20), ``trees" (20), ``flags" (20), ``calm water" (20), ``fountains" (20), ``smoke" (16), ``escalator" (7), ``traffic" (9) and ``rotation" (10), where the numbers represent the number of sequences in each class.

The \textit{DynTex gamma} dataset \cite{peteri2010dyntex} contains 264 sequences grouped into 10 classes:
``flowers" (29), ``sea" (38), ``naked trees" (25), ``foliage" (35), ``escalator" (7), ``calm water" (30), ``flags" (31), ``grass" (23), ``traffic" (9) and ``fountains" (37), where the numbers represent the number of sequences in each class.

\textit{Dyntex++} \cite{ghanem2010maximum} is derived from the DynTex database with images cropped and processed. It consists of 36 classes of 100 sequences.
Each sequence contains 50 greyscale images of size $50\times 50$. 
We reproduce the experimental setup from \cite{ghanem2010maximum,tiwari2016improved,andrearczyk2015dynamic}.
50 sequences are randomly selected from each class as the training set, and the other 50 sequences are used in testing. This
process is repeated 20 times to obtain the average classification rate.

The UCLA database \cite{doretto2003dynamic} contains 50 classes of four DT sequences each.
Each DT sequence includes 75 greyscale frames of size $160\times 110$ which are cropped to the size $48\times 48$ to show representative dynamics.
The classes can be grouped together to form more challenging sub-datasets. We reproduce the three mostly used setups which are defined as follows.

\textit{UCLA 50-class}: For this experiment, the setup is reproduced from \cite{ghanem2010maximum,xu2011dynamic,tiwari2016improved} .
A 4-fold cross-validation is performed with three sequences of each class used for training (i.e., 150 sequences), the remaining one for testing (i.e., 50 sequences) in each of the four splits.

\textit{UCLA 9-class}: For the 9-class case, we reproduce the experimental setup from \cite{ghanem2010maximum,xu2011dynamic,tiwari2016improved}.
The sequences taken from different viewpoints are grouped into 9 classes: ``boiling water" (8), ``fire" (8), ``flowers" (12), ``fountains" (20), ``plants" (108), ``sea" (12), "smoke" (4), ``water" (12) and ``waterfall" (16), where the numbers represent the number of sequences in each class.
In each class, 50\% of the sequences are used for training (i.e., 100 sequences), the other 50\% for testing (i.e., 100 sequences).
We report the average classification rate over 20 trials with random splits that are created once and used unchanged throughout the experiments.

\textit{UCLA 8-class} is similar to the 9-class setup except that the ``plant" category is removed since it contains many more sequences than the other classes.
The number of remaining sequences is 92 (i.e., 46 for training and 46 for testing).
We apply the same 20 trials as in the 9-class setup.
This experimental setup is reproduced from \cite{xu2011dynamic,tiwari2016improved}.

\subsection{Implementation details}
\label{subsec:ImpDet}
Our networks are implemented with Caffe \cite{jia2014caffe}\footnote{An implementation of our method will be provided on GitHub}.
As explained in section \ref{sec:Methods}, we use different architectures to adapt the T-CNNs to the large difference of image sizes in the various experiments.
However, we report the results without distinction as we use the same approach described in Section \ref{subsec:DTCNN} for all the experiments.

The hyperparameters of the networks slightly depend on the size of the datasets, setups and input sizes and are summarized in Table \ref{tab:hyperparam}.
\begin{table*}[!t]
\caption{Hyperparameters used for training both T-CNNs based on AlexNet and GoogleNet on different datasets.
From left to right: initial learning rate, factor gamma by which the learning rate is multiplied at every step, weight decay, momentum, batch size, number of iterations and
steps} \label{tab:hyperparam}
\centering
\resizebox{1\textwidth}{!}{
\begin{tabular}{|c|c|c|c|c|c|c|c|}
\hline
	hyper-param.            & lr        & $\gamma$  & weight decay  & momentum  & batch size    & nb. iter  & steps \\ \hline
	Dyn++                   & 0.01      & 0.01      & 0.0005        & 0.9       & 64            & 25000         & 5000,20000 \\ \hline
	UCLA-9                  & 0.01      & 0.01      & 0.0005        & 0.9       & 64            & 4000          & 1000,3000 \\ \hline
	Dyn, UCLA-50, UCLA-8    & 0.0001    & 0.1       & 0.004         & 0.9       & 64            & 2000          & 1500 \\ \hline
\end{tabular}
}
\end{table*}

The number of slices per sequence is equal in the three planes and the results are relatively robust to changes of this number.
For Dyntex++ and UCLA, we use a number of slices equal to the height, width and depth of the sequences, i.e., 50 and 48 respectively.
The sequences of the DynTex database being much larger ($352\times 288\times 250$), we use only 10 slices per plane.
We also resize all the slices after extraction to $227\times 227$.

We train our networks using stochastic gradient descent with softmax and multinomial logistic loss.

\subsection{Results}
\label{subsec:Results}
The results of our DT-CNN approach on all the datasets used in our experiments are compared with the state of the art in Table \ref{tab:results}. We include the methods that obtain the best results on each dataset, namely:
9-Plane mask Directional Number transitional Graph with SVM classifier (denoted DNG) \cite{ramirez2015spatiotemporal}, Dynamic Fractal Spectrum (denoted DFS) \cite{xu2011dynamic},
spatial Transferred ConvNet Features (denoted s-TCoF) and concatenation of spatial and temporal Transferred ConvNet Features (denoted st-TCoF) \cite{qi2016dynamic}
and Multiresolution Edge Weighted Local Structure Pattern (denoted MEWLSP) \cite{tiwari2016dynamic}.
Finally, we also compare with the LBP-TOP \cite{zhao2007dynamic} as it was tested on most of the datasets and also combines an analysis of three orthogonal planes.
The results of LBP-TOP are provided in \cite{norouznezhad2012directional} and \cite{qi2016dynamic} with the original author's implementation \cite{zhao2007dynamic} for the former and their own implementation for the latter.
\begin{table*}[!t]
\caption{Accuracy results (\%) on DT datasets of the proposed DT-CNN approaches and of the state of the art.} \label{tab:results}
\centering
\resizebox{1\textwidth}{!}{
\begin{tabular}{|c|c|c|c|c|c|c|c| }
\hline
	Methods & Dyntex++ & DynTex-alpha & DynTex-beta & DynTex-gamma & UCLA-50 & UCLA-9 & UCLA-8\\ \hline
	LBP-TOP \cite{zhao2007dynamic,norouznezhad2012directional,qi2016dynamic} & 71.2 & 96.67 & 85.8 & 84.85 & 86.1 & - & 96.8 \\ \hline
	DNG \cite{ramirez2015spatiotemporal} & 93.8 & - & - & - & - & \textbf{99.6} & \textbf{99.4} \\ \hline
	DFS \cite{xu2011dynamic} & 89.9 & - & - & - & \textbf{100} & 97.5 & 99 \\ \hline
	s-TCoF \cite{qi2016dynamic} & - & \textbf{100} & 99.38 & 95.83 & - & - & -\\ \hline
	st-TCoF \cite{qi2016dynamic} & - & 98.33 & 98.15 & 98.11 & - & - & -\\ \hline
	MEWLSP \cite{tiwari2016dynamic} & 98.48 & - & -  & -  & 96.5 & 98.55 & 98.04 \\ \hline
	DT-AlexNet & 98.18 & \textbf{100} & 99.38 & \textbf{99.62} & 99.5 & 98.05 & 98.48 \\ \hline
	DT-GoogleNet & \textbf{98.58} & \textbf{100} & \textbf{100} & \textbf{99.62} & 99.5 & 98.35 & 99.02 \\ \hline
\end{tabular}
}
\end{table*}
Our method obtains consistent high accuracy results on all the datasets and overall outperforms all other methods in the literature.
The UCLA dataset contains few training samples per class and therefore our method does not outperform the current state of the art.
However, a significant improvement is shown on the larger datasets DynTex and Dyntex++.
The consistency of high accuracy results across all the datasets shows the robustness of our approach.
The deep DT-GoogleNet approach outperforms the shallower DT-AlexNet even though more challenging datasets would be required to fully demonstrate the power of using deep architectures.
In the following, we discuss the results of each experiment in more detail.

\textit{DynTex alpha}: The classification rates of our DT-AlexNet and DT-GoogleNet are 100\% on this dataset with high inter-class variability and low intra-class variability and each sequence has at least one other sequence of the same class which is very similar.
The spatial TCoF (s-TCoF) in \cite{qi2016dynamic} also obtains a 100\% classification on this sub-dataset.

\begin{figure}[!t]
\centerline{\subfloat[misclassified sequence]{\includegraphics[width=1in]{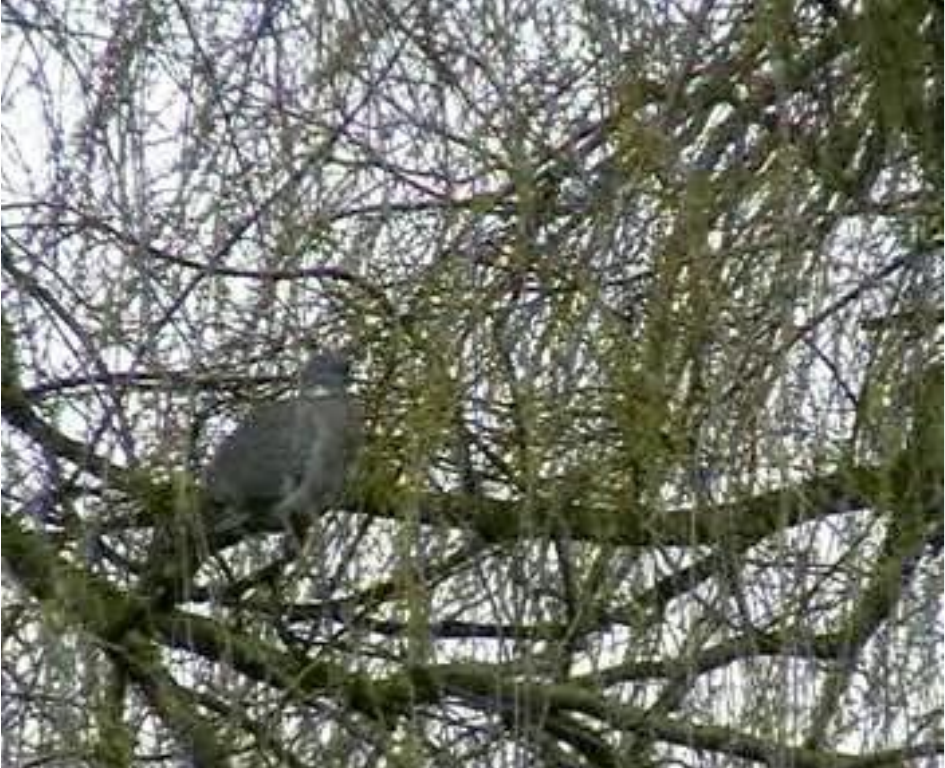}%
\label{fig:misclass_gamma_a}}
\hfil
\subfloat[true class]{\includegraphics[width=1in]{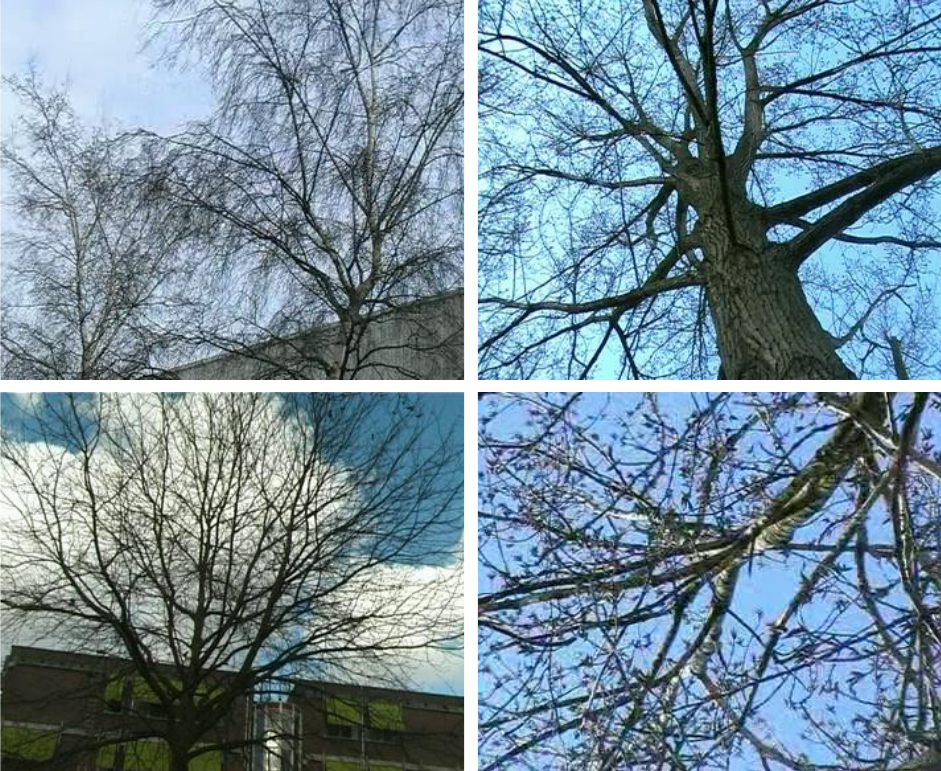}%
\label{fig:misclass_gamma_b}}
\hfil
\subfloat[detected class]{\includegraphics[width=1in]{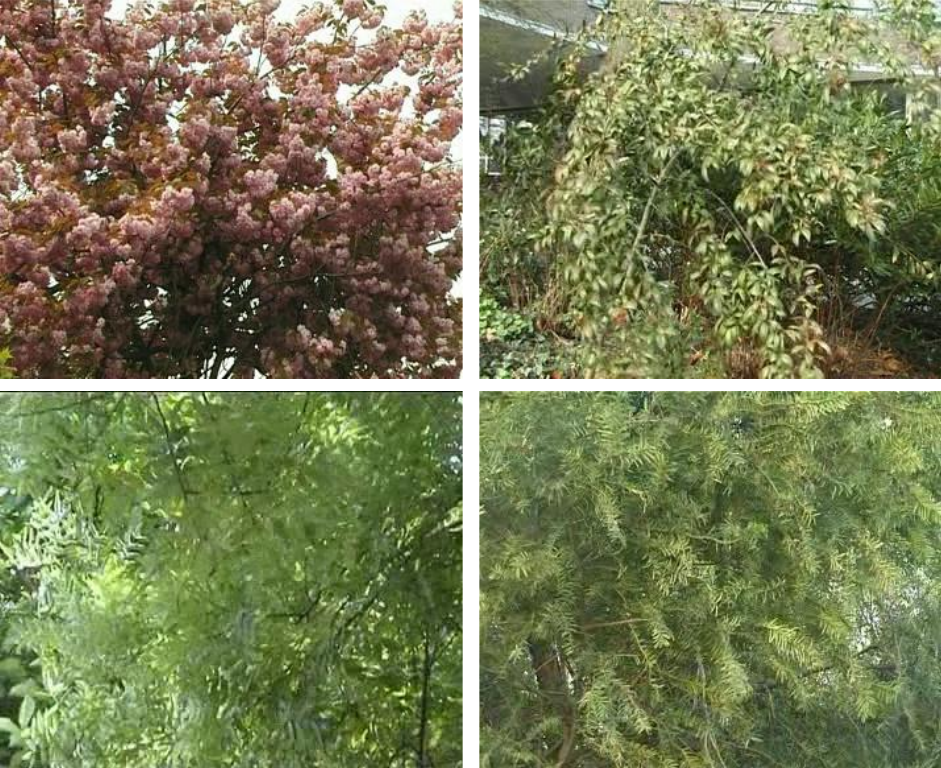}%
\label{fig:misclass_gamma_c}}}
\caption{Misclassified sequence of the DynTex gamma dataset and sequences examples from the true class and from the detected one.
(a) misclassified sequence, (b) true class ``naked trees" and (c) detected class ``foliage".}
\label{fig:misclass_gamma}
\end{figure}

\textit{DynTex beta}: The recognition rate of our DT-AlexNet is 99.38\% on DynTex-beta and 100\% with DT-GoogleNet, which improves the best result obtained by the s-TCoF \cite{qi2016dynamic}.
The only sequence misclassified with DT-AlexNet is shown in Figure \ref{fig:misclass_gamma_a} with some sequences of the true class (\ref{fig:misclass_gamma_b})
and some of the wrong detected class (\ref{fig:misclass_gamma_c}).
The background of this misclassified sequence contains mainly a blue sky and trees and occupies the major spatial part of the sequence.

\textit{DynTex gamma}: On the DynTex-gamma dataset, only one sequence is misclassified (out of 264) by our method (with both DT-CNNs),
improving the state of the art \cite{qi2016dynamic} from 98.11\% to 99.62\%.
The 79\textsuperscript{th} sequence of true class ``naked trees" is misclassified as class ``foliage" with a very similar appearance and dynamics.
As shown in Figure \ref{fig:misclass_gamma}, the tree in the misclassified sequence  is more dense as compared to other trees in the same class ``naked trees", which causes the confusion. 
This result is satisfying as DynTex-gamma contains relatively high intra-class variations (e.g., ``fountains", ``flags")
and shows inter-class similarities, e.g. ``naked trees" vs. ``foliage" and ``sea" vs. ``calm water" both in the spatial appearance and dynamics.

\begin{figure}[!t]
\centerline{\subfloat[misclassified sequence]{\includegraphics[width=1in]{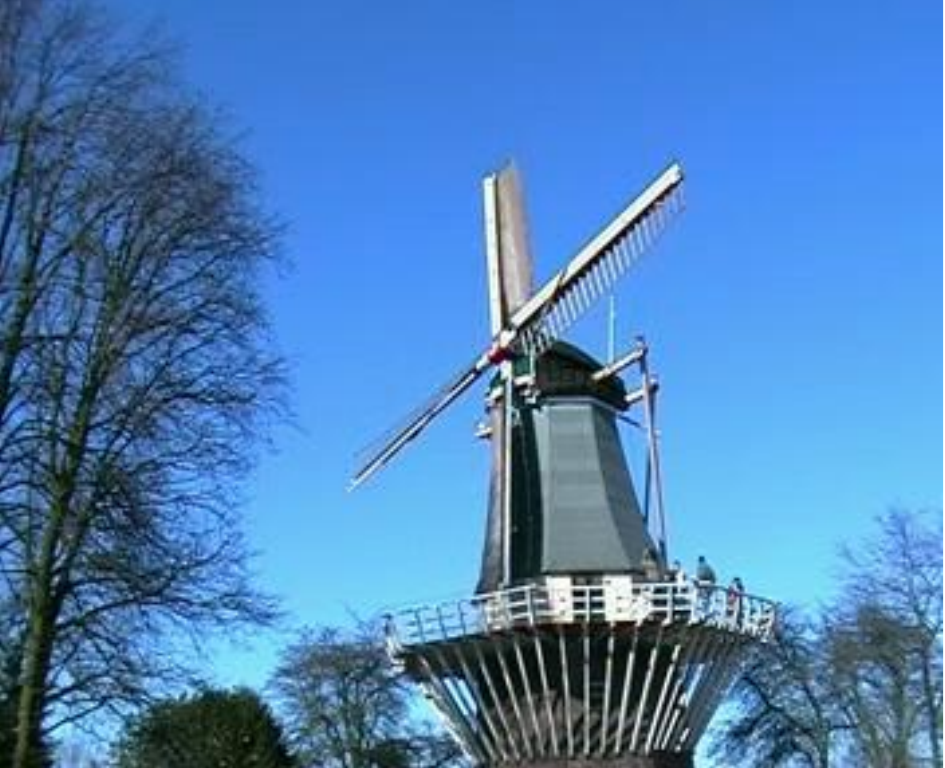}%
\label{fig:misclass_beta_a}}
\hfil
\subfloat[true class]{\includegraphics[width=1in]{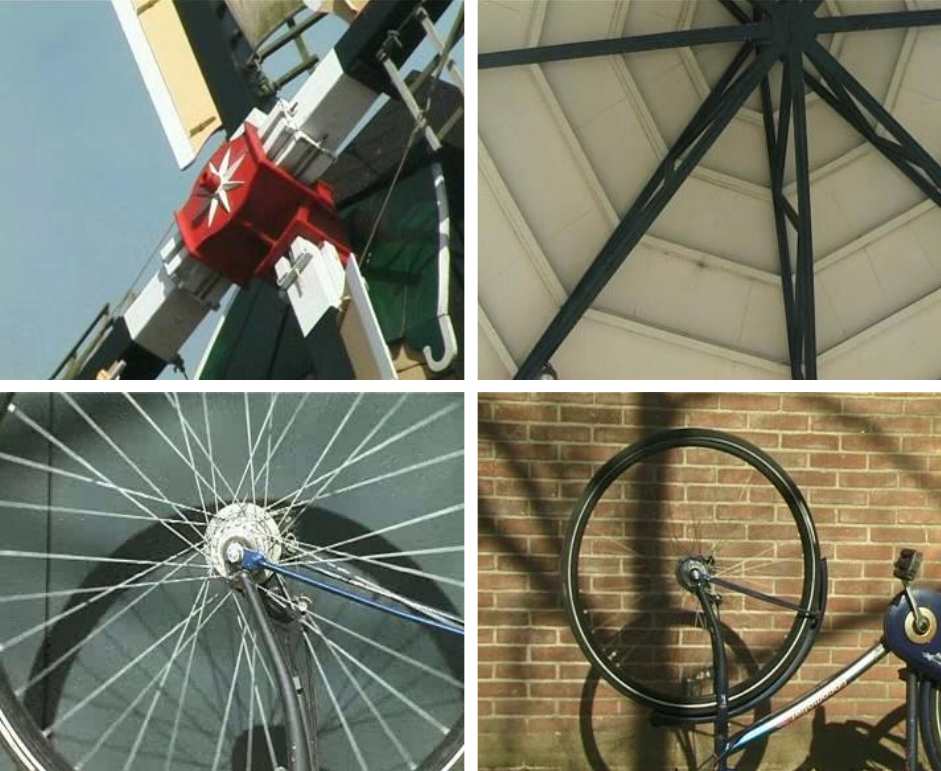}%
\label{fig:misclass_beta_b}}
\hfil
\subfloat[detected class]{\includegraphics[width=1in]{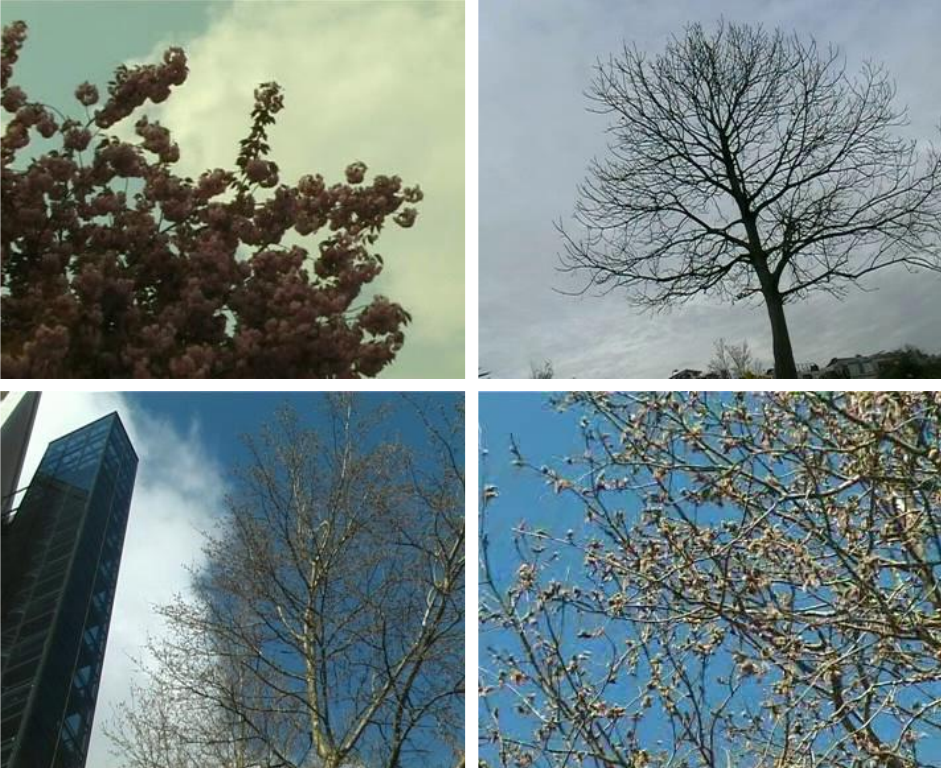}%
\label{fig:misclass_beta_c}}}
\caption{Misclassified sequence of the DynTex beta dataset and sequences examples from the true class and from the detected one.
(a) misclassified sequence, (b) true class ``rotation" and (c) detected class ``trees".}
\label{fig:misclass_beta}
\end{figure}

\begin{figure}[!t]
\centering
\includegraphics[scale=1]{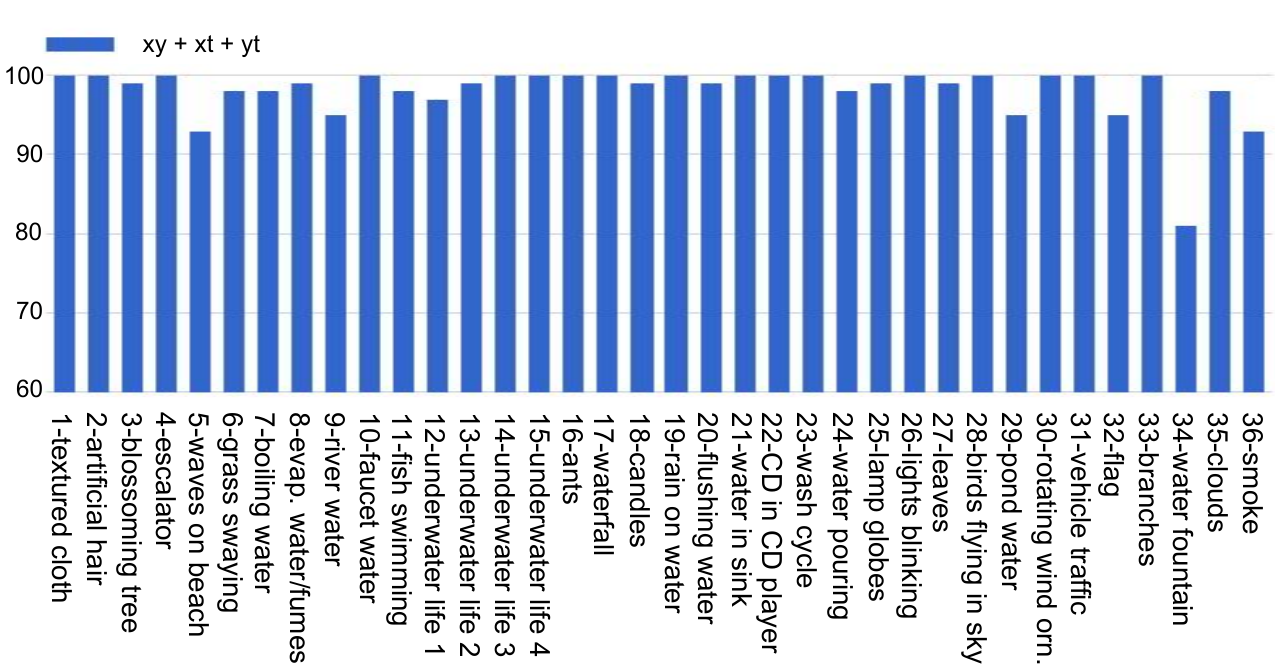}
  \caption{Classification rate of individual classes of the Dyntex++ dataset with our proposed DT-CNN approach (DT-AlexNet).}\label{fig:chart_dyntex++}
\end{figure}

\textit{Dyntex++}: In this experiment, our DT-GoogleNet approach also outperforms the best results in the literature \cite{tiwari2016dynamic}.
The classification rate of each of the 36 classes with DT-AlexNet is shown in Figure \ref{fig:chart_dyntex++}.
The most misclassifications occur for the classes with high intra-class variation ``water fountain" and ``smoke", with sequences which, with a closer look, do not always exhibit the expected DT due to the automatic splitting process of the original sequences from DynTex described in \cite{ghanem2010maximum}.
We do not illustrate the confusion matrix due to the large number of classes, yet we notice by visualizing it that there are no dominant categories with which these misclassified sequences are confused (i.e., confusions are spread over several classes). 
This experiment shows the effectiveness of our neural network approach with a relatively high number of training samples and classes as well as high intra-class variation.

The UCLA database contains only four sequences per primary category (50 classes case). Therefore the number of training sequences per primary category is three, which is much lower than the other datasets.
CNNs require a high number of training samples to adjust the weights in order to generalize the recognition task to new unknown data and avoid overfitting the training data.
The database being not highly challenging with low intra-class variation, our method (with both DT-AlexNet and DT-GoogleNet) is still able to reach nearly 100\% classification and is close to the state of the art.
However, due to the training size, we do not outperform the best shallow methods in the literature on the three sub-datasets as described below.

\textit{UCLA-50}: We achieve 99.5\% classification with only one sequence misclassified out of 200 with both DT-CNNs.
The best performance in the literature is 100\%, achieved in \cite{xu2011dynamic}, though our DT-CNNs largely outperforms this approach by over 8\% on the Dyntex++ dataset.

\textit{UCLA-9}: Our method obtains 98.05\% and 98.35\% classification rate with DT-AlexNet and DT-GoogleNet respectively,
close to the state of the art (99.6\%) \cite{ramirez2015spatiotemporal}.
The confusion matrix obtained with DT-AlexNet is shown in Table \ref{tab:confUCLA9} for which one should keep in mind that the number of sequences per class varies.
One can notice that the number of training samples of a class largely influences the recognition rate of the test samples of that same class.
The ``smoke" class contains only four samples and obtains the lowest 75\% classification.
These results confirm that our method performs best on large datasets with a high number of training samples per class due to the deep learning scheme.
\begin{table}[!t]
\caption{Confusion matrix of the proposed DT-CNN on UCLA 9-class.} \label{tab:confUCLA9}
\centering
\begin{tabular}{c*{9}{|E}|}
 \multicolumn{1}{c}{} & \multicolumn{1}{c}{boil} & \multicolumn{1}{c}{fire} 
  & \multicolumn{1}{c}{flow.} & \multicolumn{1}{c}{pl.} & \multicolumn{1}{c}{fount} & \multicolumn{1}{c}{sea} 
  & \multicolumn{1}{c}{sm.} & \multicolumn{1}{c}{wat.}  & \multicolumn{1}{c}{wfall}  \\ \hhline{~*9{|-}|}
 boil & 1 & 0 & 0 & 0 & 0 & 0 & 0 & 0 & 0 \\ \hhline{~*9{|-}|}
 fire & 0 & 0.97 & 0 & 0 & 0 & 0 & 0.03 & 0 & 0 \\ \hhline{~*9{|-}|} 
 flower & 0 & 0 & 0.92 & 0 & 0.08 & 0 & 0 & 0 & 0 \\ \hhline{~*9{|-}|}
 fount & 0 & 0 & 0 & 0.96 & 0 & 0 & 0 & 0 & 0.04 \\ \hhline{~*9{|-}|}
 plants & 0 & 0 & 0 & 0 & 1 & 0 & 0 & 0 & 0 \\ \hhline{~*9{|-}|}
 sea & 0 & 0 & 0 & 0 & 0 & 0.97 & 0 & 0.03 & 0 \\ \hhline{~*9{|-}|}
 smoke & 0 & 0 & 0 & 0 & 0 & 0 & 0.75 & 0.25 & 0 \\ \hhline{~*9{|-}|}
 water & 0 & 0 & 0 & 0 & 0 & 0 & 0.03 & 0.97 & 0 \\ \hhline{~*9{|-}|}
 wfalls & 0 & 0 & 0 & 0.01 & 0 & 0 & 0 & 0 & 0.99 \\ \hhline{~*9{|-}|}
\end{tabular}
\end{table}
\begin{table}[!t]
\caption{Confusion matrix of the proposed DT-AlexNet on UCLA 8-class.} \label{tab:confUCLA8}
\centering
\begin{tabular}{c*{8}{|E}|}
 \multicolumn{1}{c}{} & \multicolumn{1}{c}{boil} & \multicolumn{1}{c}{fire} 
  & \multicolumn{1}{c}{flower} & \multicolumn{1}{c}{fount} & \multicolumn{1}{c}{sea} 
  & \multicolumn{1}{c}{smoke} & \multicolumn{1}{c}{water}  & \multicolumn{1}{c}{wfalls}  \\ \hhline{~*8{|-}|}
 boil & 0.99 & 0.01 & 0 & 0 & 0 & 0 & 0 & 0 \\ \hhline{~*8{|-}|}
 fire & 0.01 & 0.99 & 0 & 0 & 0 & 0 & 0 & 0 \\ \hhline{~*8{|-}|} 
 flower & 0 & 0 & 1 & 0 & 0 & 0 & 0 & 0 \\ \hhline{~*8{|-}|}
 fount & 0 & 0 & 0 & 1 & 0 & 0 & 0 & 0 \\ \hhline{~*8{|-}|}
 sea & 0 & 0 & 0 & 0 & 0.95 & 0.02 & 0.02 & 0.01 \\ \hhline{~*8{|-}|}
 smoke & 0 & 0 & 0 & 0 & 0 & 1 & 0 & 0 \\ \hhline{~*8{|-}|}
 water & 0 & 0 & 0 & 0 & 0 & 0 & 1 & 0 \\ \hhline{~*8{|-}|}
 wfalls & 0 & 0 & 0 & 0.01 & 0 & 0.02 & 0.01 & 0.96 \\ \hhline{~*8{|-}|}
\end{tabular}
\end{table}

\textit{UCLA-8}: Our approach achieves a classification rate of 98.48\% with DT-AlexNet and 99.02\% with DT-GoogleNet,
close to the state of the art on this sub-dataset  of 99.4\% \cite{ramirez2015spatiotemporal}.
The confusion matrix in Table \ref{tab:confUCLA8} shows that the few confusions with DT-AlexNet mainly occur for classes exhibiting high similarities
in the spatial appearance and/or dynamics (``sea", ``smoke", ``water" and ``waterfalls").

\begin{table*}[!t]
\caption{Accuracy results (\%) on DT datasets of the proposed DT-AlexNet using various combinations of planes.} \label{tab:results_planes}
\centering
\resizebox{1\textwidth}{!}{
\begin{tabular}{ |c|c|c|c|c|c|c|c| }
\hline
	Methods & Dyntex++ & DynTex-alpha & DynTex-beta & DynTex-gamma & UCLA-50 & UCLA-9 & UCLA-8\\ \hline
	$xy$ & 94.28 & \textbf{100} & 98.77 & 98.11 & 99 & 95.7 & 98.04 \\ \hline
	$xt$ & 96.57 & \textbf{100} & 95.68 & 97.35 & 98.5 & 97.4 & 95.76 \\ \hline
	$yt$ & 96.28 & \textbf{100} & 95.06 & 97.73 & 93 & 97.15 & 95.65 \\ \hline
	$xy+xt$ & 97.57 & \textbf{100} & \textbf{99.38} & 99.24 & \textbf{99.5} & 97.4 & 98.26 \\ \hline
	$xy+yt$ & 97.71 & \textbf{100} & \textbf{99.38} & \textbf{99.62} & 99 & 97.8 & \textbf{98.59} \\ \hline
	$xt+yt$ & 97.84 & \textbf{100} & 97.53 & 98.11 & 98 & 97.95 & 96.85 \\ \hline
	$xy+xt+yt$ & \textbf{98.18} & \textbf{100} & \textbf{99.38} & \textbf{99.62} & \textbf{99.5} & \textbf{98.05} & 98.48 \\ \hline
\end{tabular}
}
\end{table*}

\subsection{Contribution of planes}
The results obtained with DT-AlexNet using different combinations of planes are reported in Table \ref{tab:results_planes}.
\subsubsection{Single plane analysis}\leavevmode \par
Firstly, while the spatial analysis of the $xy$ plane is better overall than the temporal ones ($xt$ and $yt$), the temporal analyses are close to, and at times outperform, the spatial analysis.

Note that the performance of the $xt$ and $yt$ planes for some classes depends a lot on the angle at which the DT videos are captured as high temporal information is sometimes captured in one or the other plane.
For instance, translational motions such as waves and traffic may be contained in only one of the temporal planes if the direction of the motion is parallel/perpendicular to one of the spatial axes.
Furthermore, an independent CNN is fine-tuned on each plane and thus learns differently.
Therefore, we notice differences between the temporal analyses of the $xt$ and $yt$ planes.

The good results obtained with both single temporal planes go against early conclusions made in the literature stating that DT recognition mostly relies on the spatial analysis while the motion analysis can only provide minor complementary information \cite{ghanem2010maximum,ren2013dynamic,andrearczyk2015dynamic}.
This statement is based on experimental results and on the human perception of DTs.
While it might be true for the experiments conducted in \cite{ghanem2010maximum,ren2013dynamic,andrearczyk2015dynamic} and for a human being able to differentiate most DTs with a single frame,
we show that it does not generalize to all the analysis methods, and that neural network approaches are excellent at finding their own ways to analyze things.
Indeed, when it might be very difficult for a human to recognize sequences of temporal slices like those shown in Figures \ref{fig:slicesxt} and \ref{fig:slicesyt}, our approach is able to accurately classify sequences given only the temporal slices. 

Similar observations were made in \cite{zhao2007dynamic} in which the LBP histograms on all three single planes result in similar accuracies on the DynTex dataset.
One should note that in both our approach and \cite{zhao2007dynamic}, the analysis of the $xt$ and $yt$ planes are not purely temporal as they reflect the evolution of 1D spatial lines of pixels over time.
Yet these planes exhibit temporal variations and these types of approaches are generally referred to as temporal analysis.

\begin{figure}[!t]
\centerline{\subfloat[ucla-8]{\includegraphics[scale=.8]{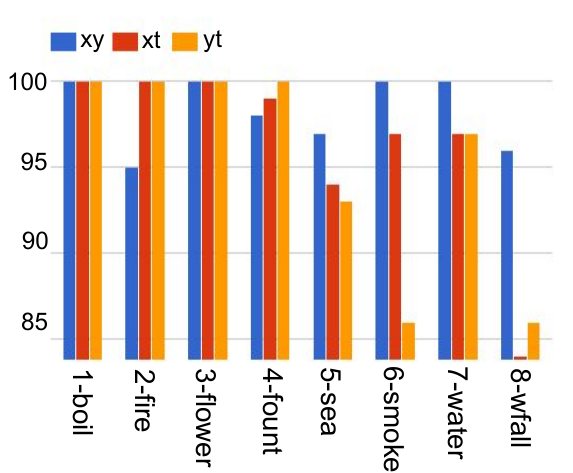}%
\label{fig:chart_ucla8_xy_xt_yt}}
\hfil
\subfloat[ucla-9]{\includegraphics[scale=.8]{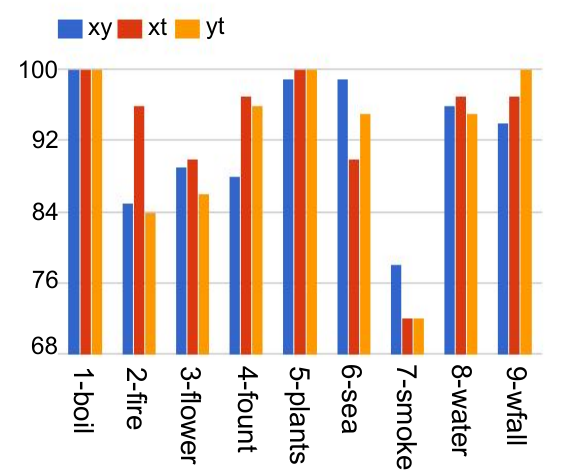}%
\label{fig:chart_ucla9_xy_xt_yt}}}
\caption{Classification rate of individual classes using single $xy$, $xt$ and $yt$ planes with DT-AlexNet on the (a) ucla-8 and (b) ucla-9 sub-datasets.}
\label{fig:misclass_gamma}
\end{figure}

\begin{figure}[!t]
\centerline{\subfloat[DynTex beta]{\includegraphics[scale=.8]{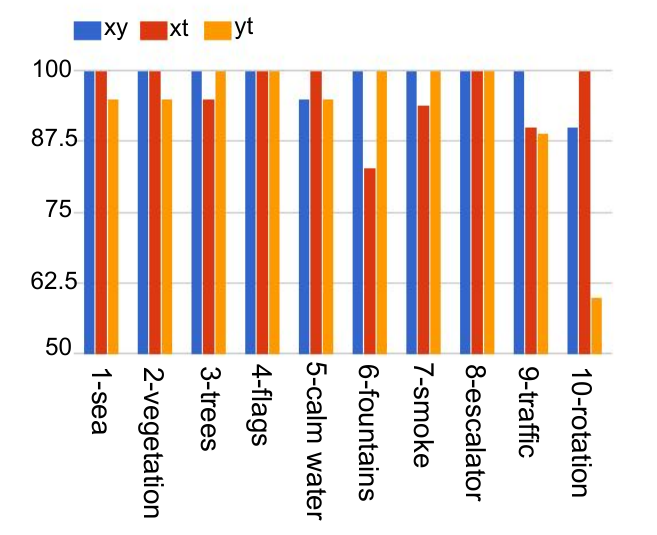}%
\label{fig:chart_dyntexB_xy_xt_yt}}
\hfil
\subfloat[DynTex gamma]{\includegraphics[scale=.8]{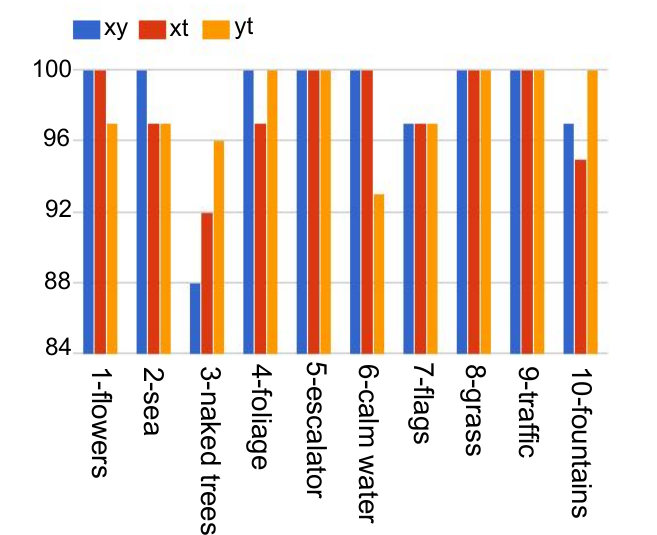}%
\label{fig:chart_dyntexC_xy_xt_yt}}}
\caption{Classification rate of individual classes using single $xy$, $xt$ and $yt$ planes with DT-AlexNet on the (a) DynTex beta and (b) DynTex gamma sub-datasets.}
\label{fig:misclass_gamma}
\end{figure}

\begin{figure}[!t]
\centering
\includegraphics[scale=1]{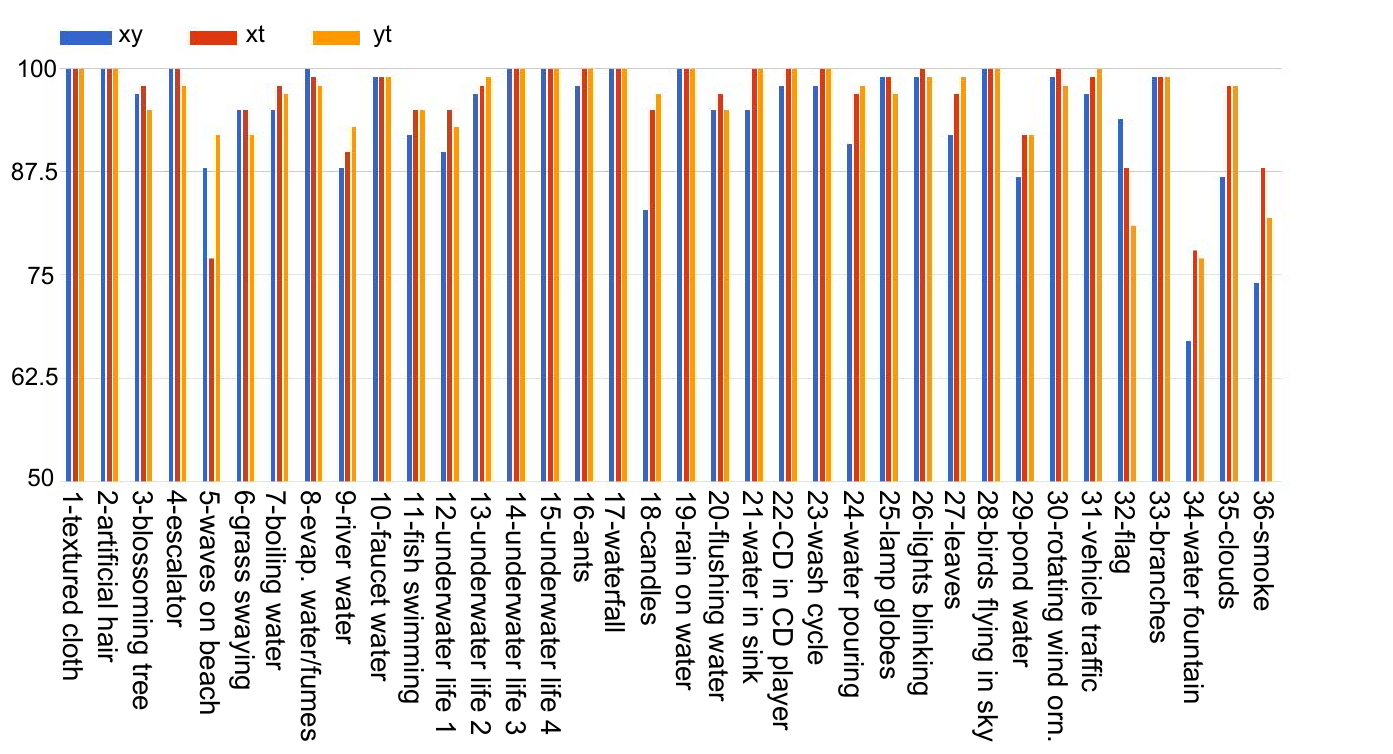}
  \caption{Classification rate of individual classes of the Dyntex++ dataset using single $xy$, $xt$ and $yt$ planes with DT-AlexNet.}\label{fig:chart_dyntex++_xy_xt_yt}
\end{figure}

In our experiments, the spatial analysis ($xy$ plane) performs 1\% to 3\% better than the temporal ones ($xt$ and $yt$) for most of the sub-datasets (DynTex beta, DynTex gamma, UCLA-50 and UCLA-8).
However, the temporal analysis outperforms the spatial one on the Dyntex++ and the UCLA-9 sub-datasets.
The accuracy of each class using single plane analysis is shown in Figures \ref{fig:chart_ucla8_xy_xt_yt}, \ref{fig:chart_ucla9_xy_xt_yt}, \ref{fig:chart_dyntexB_xy_xt_yt}, \ref{fig:chart_dyntexC_xy_xt_yt} and \ref{fig:chart_dyntex++_xy_xt_yt} for respectively the UCLA-8, UCLA-9, DynTex-beta, DynTex-gamma and Dyntex++ datasets.
It is difficult to find consistent results across datasets to conclude what recognition the spatial analysis is able to achieve better than the temporal ones and/or to attempt to generalize these observations.


\subsubsection{Combination of planes}\leavevmode \par
The combination of all the planes results in a consistent increase of accuracy as compared to the sole spatial analysis ($xy$ plane).
Similarly to \cite{zhao2007dynamic}, these results show the complementarity of the spatial and temporal planes.

In \cite{zhao2007dynamic}, the authors assign weights to the planes depending on the test results in order to increase the recognition rate.
We want to keep a fully automated system and prefer not to report such experiments. 
Also, we do not assign weights based on the training loss which varies drastically or another measure of performance calculated on the training set because the learned parameters do not always generalize well to the test set; e.g., a low training loss may result in low test accuracy if the network overfits the training data.
Finally, we do not split the training data to learn which planes perform best on unknown data because, firstly, we want to keep the same setup as in the literature; secondly, we want to maximize the number of training data for training the CNNs.

The $xy$ plane captures the spatial information while the other two planes mainly capture the dynamics of the sequences.
As expected, combining the spatial plane $xy$ to a single temporal plane, either $xt$ or $yt$ performs generally better than the sole $xy$ plane or the two temporal planes combined.
The best overall performance is achieved by the combination of the three planes which again shows their complementarity and the ability of our method to combine them in a collective classification.

\subsection{Domain transferability and visualization}\leavevmode \par
\begin{figure*}[!t]
\centerline{\subfloat[xy+xt+yt]{\includegraphics[scale=.8]{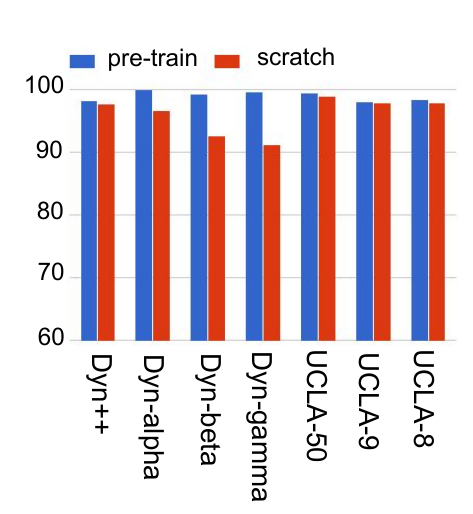}%
\label{fig:chart_scratch_xyt}}
\hfil
\subfloat[xy]{\includegraphics[scale=.8]{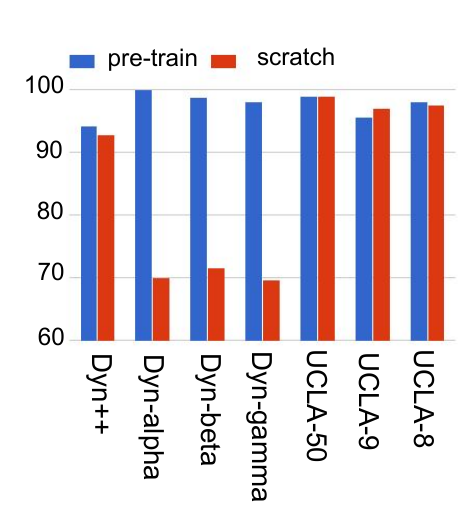}%
\label{fig:chart_scratch_xy}}
\hfil
\subfloat[xt]{\includegraphics[scale=.8]{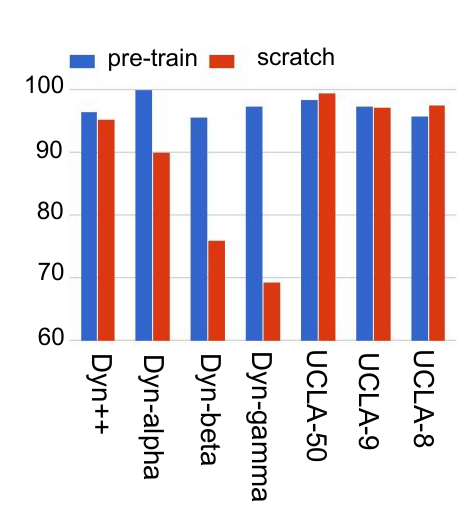}%
\label{fig:chart_scratch_xt}}
\hfil
\subfloat[yt]{\includegraphics[scale=.8]{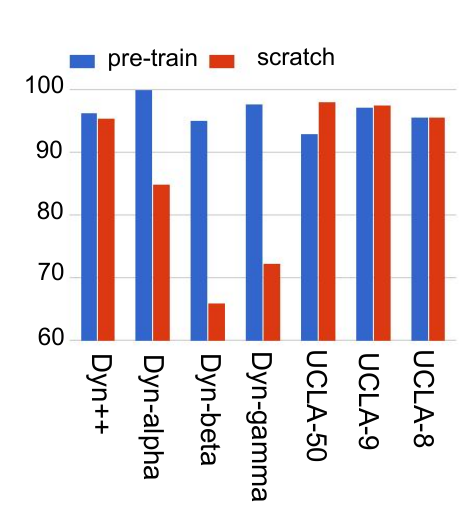}%
\label{fig:chart_scratch_yt}}}
\caption{Classification rate of DT-AlexNet with networks trained from scratch vs pre-trained on ImageNet with the following planes: (a) $xy$+$xt$+$yt$ (b) xy, (c) xt and (d) yt.}
\label{fig:chart_scratch}
\end{figure*}

In this section, we will start with a comparison of DT-AlexNet using networks from scratch and pre-trained. The results are compared in Figure \ref{fig:chart_scratch}.
We notice that using pre-trained networks only slightly improves the accuracy of our method using the combination of three plans (see Figure \ref{fig:chart_scratch_xyt}).
The networks are able to learn from scratch due to the relatively large number of samples resulting from the slicing approach.

As shown in Figure \ref{fig:chart_scratch}, the networks used for Dyntex++ and UCLA (T-CNN50) learn better from scratch than the larger T-CNNs used for the DynTex datasets.
It is particularly true for single plane methods (Figures \ref{fig:chart_scratch_xy} to \ref{fig:chart_scratch_yt}).
Having less parameters, they learn better from small datasets with less overfitting and thus do not benefit as much from the pre-training.

Moreover, one could expect the spatial analysis to benefit extensively more from the pre-training than the temporal analysis
as the source (ImageNet) and target (spatial textures) domains are closer.
Yet, the temporal planes benefit almost similarly to the spatial one which demonstrates
the domain transferability of the learned parameters across domains.
Note that it also shows that the learning is not biased by an overlap (similar images and classes) between the pre-training ImageNet dataset and the  
DT datasets.

We will now comment on the kernels learned by our method and the features that they detect.
First of all, we notice that in ImageNet, some classes contain images with high texture content (tone variation and repetitive patterns) such as ``cheetah", ``chainlink fence", ``theater curtains" and ``grille".
As shown in \cite{andrearczyk2016using}, pre-training  T-CNN with such classes transfers better on texture datasets than pre-training with object-like classes.
We also notice here that inputting spatial or temporal slices from the DynTex database to T-CNN pre-trained on ImageNet often highly activates neurons that learned to recognize these texture classes.
Thus the network is able to learn texture-like patterns already from non segmented textures (e.g. cheetas) and this is why it transfers well to (spatial and temporal) texture images.
Moreover, the response at the last convolution layer of the T-CNN to both spatial and temporal DT slices is more dense than the response to an object image.
This is also expected and motivates the use of the energy layer (average pooling) for texture analysis and DT analysis.

We also compare the features that the neurons of the fine-tuned and scratch T-CNNs respond most to using the image synthesis to produce high activation of neurons via gradient ascent optimization from \cite{yosinski2015understanding}.
We notice, as one would expect, that the neurons of the intermediate layers respond to more complex patterns in fine-tuned networks than those trained from scratch.
However, there is very little difference in complexity between the maximum activation of the neurons of the last fully-connected layer. The neuron mainly detect features that exhibit simple repeated patterns.
It shows that the dense orderless pooling strategy of the energy layer discards the overall shape information to focus on the repetition of more simple patterns.

Finally, note that for some datasets, the temporal slices $xt$ and $yt$ slices could be probabilistically expected to exhibit the same dynamic.
It would be the case for instance with random rotations of the field of view across different sequences and/or for DT sequences without dominant orientation of motion.
In such scenario, it could be useful to combine the temporal slices $xt$ and $yt$ into a single analysis i.e. using two fine-tuned networks instead of three and obtain more rotation invariance.
This approach did not result in an increase of accuracy in the experiments proposed in this paper.

\section{Conclusion}
In this paper, we developed a new approach designed for the analysis of DT sequences based on CNNs applied on three orthogonal planes.
We  have shown that training independent CNNs on three orthogonal planes and combining their outputs in an ensemble model
manner performs well on DT classification by learning to jointly recognize spatial and dynamic patterns.
We based this work on our previously described T-CNN (specifically designed for texture images) and developed a new network for
small images (Dyntex++ and UCLA) as well as deeper T-CNN networks adapted from GoogleNet to texture analysis.
We experimented with our approach on the most used DT datasets in the literature,
yet a major problem remains the lack of larger and more challenging DT datasets to fully exploit the power of deep learning.
Our deepest approach (DT-GoogleNet) obtains slightly higher accuracy than the shallower one (DT-AlexNet) and established a new state of the art on
the DynTex and Dyntex++ databases.
It achieved less than 1\% lower accuracy than the state of the art on the UCLA database which contains fewer training samples,
making the CNN training difficult.
The neural network learning process enables our approach to analyze and learn from many diverse datasets and setups with a good invariance to rotation,
illumination, scale, sequence length and camera motion.
Thus our method obtains high accuracy in all the tested datasets ($>$98\%) whereas previous methods in the literature are more specialized in the
analysis of one particular database.

We also showed high accuracy results using single temporal planes $xt$ and $yt$, at times outperforming the spatial analysis of the $xy$ plane.
It demonstrates both the domain transferability of the features learned only on spatial images and the importance of temporal analysis in DT recognition.
Finally, we demonstrated the complementarity of the spatial and temporal analyses with the best results obtained by the combination of the three planes
over single and two planes analyses. 

\section{Future research}
Our DT-CNN approach can be used with any network architecture, yet little difference is observed between shallow handcrafted methods and
deep models due to the saturation of the DT datasets.
A more challenging and larger DT database is required to demonstrate the advantage of deep models over shallow ones.

Finally, an interesting line of research would be to incorporate the analysis of the three planes into a single network architecture
with an early or slow fusion instead of the late fusion adopted here or by allowing connections in intermediate layers.
\small{
\bibliographystyle{ieeetr}
\bibliography{bibli}
}

\end{document}